\documentclass[twoside,11pt]{article}

\usepackage{pdflscape}
\usepackage{booktabs}
\usepackage{multirow}
\usepackage{paralist,amsmath, amssymb, bm}
\usepackage{algorithm,algorithmic}
\usepackage{jmlr2e}

\usepackage[colorlinks,
            linkcolor=blue,
            citecolor=blue,
            urlcolor=magenta,
            linktocpage,
            plainpages=false]{hyperref}

\makeatletter
\newcounter{ALC@tempcntr}
\makeatother

\makeatletter
\AtBeginDocument{\Hy@breaklinkstrue}
\makeatother

\def \y {\mathbf{y}}

\def \x {\mathbf{x}}

\def \u {\mathbf{u}}

\def \w {\mathbf{w}}
\def \R {\mathbb{R}}

\def \N {\mathbb{N}}
\def \A {\mathcal{A}}

\def \I {\mathcal{I}}

\def \wh {\widehat{\w}}

\def \S {\mathcal{S}}

\newtheorem{ass}{Assumption}
\newtheorem{thm}{Theorem}

\DeclareMathOperator*{\Reg}{Regret}
\DeclareMathOperator*{\WAReg}{WA-Regret}
\DeclareMathOperator*{\SAReg}{SA-Regret}

\DeclareMathOperator*{\argmin}{argmin}

\begin{document}

\title{Dual Adaptivity: A Universal Algorithm for Minimizing the Adaptive Regret of Convex Functions}

\author{\name Lijun Zhang \email zhanglj@lamda.nju.edu.cn\\
       \name Guanghui Wang \email wanggh@lamda.nju.edu.cn\\
       \addr National Key Laboratory for Novel Software Technology, Nanjing University, Nanjing 210023, China
       \AND
       \name Wei-Wei Tu \email tuwwcn@gmail.com\\
       \addr 4Paradigm Inc., Beijing, China
       \AND
       \name Zhi-Hua Zhou \email zhouzh@lamda.nju.edu.cn\\
       \addr National Key Laboratory for Novel Software Technology, Nanjing University, Nanjing 210023, China}
       
\editor{}

\maketitle

\begin{abstract}
To deal with changing environments, a new performance measure---adaptive regret, defined as the maximum static regret over any interval, was  proposed in online learning. Under the setting of online convex optimization, several algorithms have been successfully developed to minimize the adaptive regret. However, existing algorithms lack universality in the sense that they can only handle one type of convex functions and need apriori knowledge of parameters. By contrast, there exist universal algorithms, such as MetaGrad, that attain optimal static regret for multiple types of convex functions simultaneously. Along this line of research, this paper presents the \emph{first} universal algorithm for minimizing the adaptive regret of convex functions. Specifically, we borrow the idea of maintaining multiple learning rates in MetaGrad to handle the uncertainty of functions, and utilize the technique of sleeping experts to capture changing environments. In this way, our algorithm automatically adapts to the property of functions (convex, exponentially concave, or strongly convex), as well as the nature of environments (stationary or changing). As a by product, it also allows the type of functions to switch between rounds.
\end{abstract}

\section{Introduction}
Online learning aims to make a sequence of accurate decisions given knowledge of answers to previous tasks and possibly additional information \citep{Online:suvery}. It is performed in a sequence of consecutive rounds,
where at round $t$ the learner is asked to select a decision $\w_t$ from a domain $\Omega$. After submitting the answer, a loss function $f_t:\Omega \mapsto \R$ is revealed and the learner suffers a loss $f_t(\w_t)$.  The standard performance measure is the regret \citep{bianchi-2006-prediction}:
\[
\Reg(T)=\sum_{t=1}^T f_t(\w_t) - \min_{\w \in \Omega} \sum_{t=1}^T f_t(\w)
\]
defined as the difference between the cumulative loss of the online learner and that of the best decision chosen in hindsight. When both the domain $\Omega$ and the loss $f_t(\cdot)$ are convex, it becomes online convex optimization (OCO) \citep{zinkevich-2003-online}.

In the literature, there exist  plenty of algorithms to minimize the regret under the setting of OCO  \citep{Intro:Online:Convex}. However, when the environment undergoes many changes, regret may not be the best measure of performance. That is because regret chooses a fixed comparator, and for the same reason, it is also referred to as \emph{static} regret. To address this limitation, \citet{Adaptive:Hazan} introduce the concept of adaptive regret, which measures the performance with respect to a changing comparator. Following the terminology of \citet{Adaptive:ICML:15}, we define the strongly adaptive regret as the maximum static regret over intervals of length $\tau$, i.e.,
\begin{equation} \label{eqn:strong:adaptive}
\SAReg(T,\tau) = \max_{[p, p+\tau -1] \subseteq [T]} \left(\sum_{t=p}^{p+\tau -1}  f_t(\w_t)  - \min_{\w \in \Omega} \sum_{t=p}^{p+\tau -1} f_t(\w) \right).
\end{equation}

Since the seminal work of \citet{Adaptive:Hazan}, several algorithms have been successfully developed to minimize the adaptive regret of convex functions, including general convex, exponentially concave (abbr.~exp-concave) and strongly convex functions \citep{Hazan:2009:ELA,Improved:Strongly:Adaptive,Dynamic:Regret:Adaptive}. However, existing methods can only handle one type of convex functions. Furthermore, when facing exp-concave and strongly convex functions, they need to know the moduli of exp-concavity and strong convexity.  The lack of universality hinders their applications to real-world problems.

One the other hand, there do exist universal algorithms, such as MetaGrad \citep{NIPS2016_6268}, that attain optimal static regret for multiple types of convex functions simultaneously. This observation motivates us to ask whether it is possible to design a single algorithm to minimize the adaptive regret of multiple types of functions. This is very challenging because the algorithm needs to enjoy dual adaptivity, adaptive to the function type and adaptive to the environment. In this paper, we provide an affirmative answer by developing a Universal algorithm for Minimizing the Adaptive regret (UMA). First, inspired by MetaGrad, UMA maintains multiple learning rates to handle the uncertainty of functions. In this way, it supports multiple types of functions simultaneously and identifies the best learning rate automatically. Second, following existing studies on adaptive regret, UMA deploys sleeping experts \citep{Freund:1997:UCP} to minimize the regret over any interval, and thus achieves a small adaptive regret and captures the changing environment.

The main advantage of UMA is that it  attains second-order regret bounds over any interval. As a result, it can minimize the adaptive regret of general convex functions, and automatically take advantage of easier functions whenever possible. Specifically, UMA attains $O(\sqrt{\tau \log T})$, $O(\frac{d}{\alpha}\log \tau \log T)$ and $O(\frac{1}{\lambda}\log \tau \log T)$ strongly adaptive regrets for general convex, $\alpha$-exp-concave and $\lambda$-strongly convex functions respectively, where $d$ is the dimensionality. All of these bounds match the state-of-the-art results on adaptive regret \citep{Improved:Strongly:Adaptive,Dynamic:Regret:Adaptive} exactly. Furthermore, UMA can also handle the case that the type of functions changes between rounds. For example, suppose the online functions are general convex during interval $I_1$, then become $\alpha$-exp-concave in  $I_2$, and finally switch to  $\lambda$-strongly convex in $I_3$. When facing this function sequence, UMA  achieves $O(\sqrt{|I_1|\log T})$, $O(\frac{d}{\alpha}\log |I_2| \log T)$ and $O(\frac{1}{\lambda}\log |I_3| \log T)$ regrets over intervals $I_1$, $I_2$ and $I_3$, respectively.
\section{Related work}
We briefly review related work on static regret and adaptive regret, under the setting of OCO.
\subsection{Static regret}
To minimize the static regret of general convex functions, online gradient descent (OGD) with step size $\eta_t=O(1/\sqrt{t})$ achieves an  $O(\sqrt{T})$  bound \citep{zinkevich-2003-online}. If all the online functions are $\lambda$-strongly convex, OGD with step size $\eta_t=O(1/[\lambda t])$ attains an $O(\frac{1}{\lambda} \log T)$ bound \citep{ICML_Pegasos}. When the functions are $\alpha$-exp-concave, online Newton step (ONS), with knowledge of $\alpha$, enjoys an $O(\frac{d}{\alpha} \log T)$ bound, where $d$ is the dimensionality \citep{ML:Hazan:2007}.  These regret bounds are minimax optimal for the corresponding type of functions \citep{Minimax:Online}, but choosing the optimal algorithm for a specific problem requires domain knowledge.

The study of universal algorithms for OCO stems from the adaptive online gradient descent (AOGD) \citep{NIPS2007_3319} and its proximal extension \citep{icml2009_033}. The key idea of AOGD is to add a quadratic regularization term to the loss.
\citet{NIPS2007_3319} demonstrate that AOGD is able to interpolate between the $O(\sqrt{T})$  bound of general convex functions and the $O(\log T)$ bound of strongly convex functions. Furthermore, it allows the online function to switch between general convex and strongly convex. However, AOGD has two restrictions:
\begin{compactitem}
  \item It needs to calculate the modulus of strong convexity on the fly, which is a nontrivial task;
  \item It does not support exp-concave functions explicitly, and thus can only achieve a suboptimal $O(\sqrt{T})$ regret for this type of functions.
\end{compactitem}

Another milestone is the multiple eta gradient algorithm (MetaGrad) \citep{NIPS2016_6268,pmlr-v99-mhammedi19a}, which adapts to a much broader class of functions, including convex and exp-concave functions. MetaGrad's main feature is that it simultaneously considers multiple learning rates  and  does not need to know the modulus of exp-concavity.
MetaGrad achieves $O(\sqrt{T \log\log T})$ and $O(\frac{d}{\alpha} \log T)$ regret bounds for general convex and $\alpha$-exp-concave functions, respectively. However, it  suffers the following two limitations:
\begin{compactitem}
  \item MetaGrad treats strongly convex functions as exp-concave, and thus only gives a suboptimal $O(\frac{d}{\lambda} \log T)$ regret for $\lambda$-strongly convex functions;
  \item It assumes the type of online functions, as well as the associated parameter, does not change between rounds.
\end{compactitem}
The first limitation of MetaGrad has been addressed by \citet{Adaptive:Maler}, who develop a universal algorithm named as multiple sub-algorithms and learning rates (Maler). It attains $O(\sqrt{T})$, $O(\frac{d}{\alpha} \log T)$ and $O(\frac{1}{\lambda} \log T)$ regret bounds for general convex, $\alpha$-exp-concave, and $\lambda$-strongly convex functions, respectively. Furthermore, \citet{AAAI:2020:Wang} extend Maler to make use of smoothness. However, the second limitation remains there.
\subsection{Adaptive regret}
Adaptive regret has been studied in the setting of prediction with expert advice  \citep{LITTLESTONE1994212,Freund:1997:UCP,Adamskiy2012,Track_Large_Expert,pmlr-v40-Luo15} and OCO. In this section, we focus on the related work in the latter one.

Adaptive regret is  introduced by \citet{Adaptive:Hazan}, and later refined  by \citet{Adaptive:ICML:15}. 
We refer to the definition of  \citeauthor{Adaptive:Hazan} as weakly adaptive regret:
\[
\WAReg(T)=  \max_{[p, q] \subseteq [T]} \left(\sum_{t=p}^{q} f_t(\w_t) - \min_{\w \in \Omega} \sum_{t=p}^{q} f_t(\w)\right).
\]
For $\alpha$-exp-concave functions, \citet{Adaptive:Hazan} propose an adaptive algorithm named as Follow-the-Leading-History (FLH). FLH restarts a copy of ONS in each round as an expert, and chooses the best one using expert-tracking algorithms. The meta-algorithm used to track the best expert is inspired by the Fixed-Share algorithm \citep{Herbster1998}.
While FLH is equipped with an $O(\frac{d}{\alpha} \log T)$ weakly adaptive regret, it is computationally expensive since it needs to maintain $t$ experts in the $t$-th iteration. To reduce the computational cost,  \citet{Adaptive:Hazan} further prune the number of experts based on a data streaming algorithm. In this way, FLH only keeps $O(\log t)$ experts, at the price of an $O(\frac{d}{\alpha} \log^2 T)$ weakly adaptive regret. Notice that the efficient version of FLH essentially creates and removes experts dynamically. As pointed out by \citet{Adamskiy2012}, this behavior can be modeled by the sleeping expert setting \citep{Freund:1997:UCP}, in which the expert can be ``asleep'' for certain rounds and does not make any advice.

For general convex functions, we can use OGD as the expert-algorithm in FLH. \citet{Adaptive:Hazan} prove that FLH and its efficient variant attain $O(\sqrt{T \log T})$ and $O(\sqrt{T \log^3 T})$ weakly adaptive regrets, respectively. This result reveals a limitation of weakly adaptive regret---it does not respect short intervals well. For example, the $O(\sqrt{T \log T})$ regret bound is meaningless for intervals of length $O(\sqrt{T})$. To address this limitation, \citet{Adaptive:ICML:15} introduce the strongly adaptive regret  which takes the interval length as a parameter, as shown in (\ref{eqn:strong:adaptive}). They propose a novel meta-algorithm, named as Strongly Adaptive Online Learner (SAOL). SAOL carefully constructs a set of intervals, then runs an instance of low-regret algorithm in each interval as an expert, and finally combines active experts' outputs by a variant of multiplicative weights method \citep{v008a006}. SAOL also maintains $O(\log t)$ experts in the $t$-th round, and achieves an $O( \sqrt{\tau} \log T )$ strongly adaptive regret for convex functions.  Later, \citet{Improved:Strongly:Adaptive} develop a new meta-algorithm named as sleeping coin betting (CB), and improve the strongly adaptive regret bound to $O(\sqrt{\tau \log T})$. Very recently, \citet{pmlr-v119-cutkosky20a} proposes a novel strongly adaptive method, which can guarantee the $O(\sqrt{\tau \log T})$ bound in the worst case, while achieving tighter results when the square norms of gradients are small.

For $\lambda$-strongly convex functions,  \citet{Dynamic:Regret:Adaptive} point out that we can replace ONS with OGD, and obtain an $O(\frac{1}{\lambda} \log T)$ weakly adaptive regret. They also demonstrate that the number of active experts can be reduced from $t$ to $O(\log t)$, at a cost of an additional $\log T$ factor in the regret. All the aforementioned adaptive algorithms need to query the gradient of the loss function at least $\Theta(\log t)$ times in the $t$-th iteration. Based on surrogate losses, \citet{Adaptive:One:Gradient} show that the number of gradient evaluations per round can be reduced to $1$ without affecting the performance.

\section{Main results}
We first present necessary preliminaries, including assumptions, definitions and the technical challenge,
then provide our universal algorithm and its theoretical guarantee.

\subsection{Preliminaries}
We introduce two common assumptions used in the study of OCO \citep{Intro:Online:Convex}.

\begin{ass}\label{ass:1} The diameter of the domain $\Omega$ is bounded by $D$, i.e.,
\begin{equation}\label{eqn:domain}
\max_{\x, \y \in \Omega} \|\x -\y\| \leq D.
\end{equation}
\end{ass}
\begin{ass}\label{ass:2} The gradients of all the online functions are bounded by $G$, i.e.,
\begin{equation}\label{eqn:gradient}
\max_{\w \in \Omega}\|\nabla f_t(\w)\| \leq G, \ \forall t \in[T].
\end{equation}
\end{ass}

Next, we state definitions of strong convexity and exp-concavity \citep{Convex-Optimization,bianchi-2006-prediction}.
\begin{definition} \label{def:strong} A function $f: \Omega \mapsto \R$ is $\lambda$-strongly convex if
\[
f(\y) \geq f(\x) +  \langle \nabla f(\x), \y -\x  \rangle + \frac{\lambda}{2} \|\y -\x \|^2,  \  \forall \x, \y \in \Omega.
\]
\end{definition}
\begin{definition} \label{def:exp} A function $f: \Omega \mapsto \R$ is $\alpha$-exp-concave if $\exp(-\alpha f(\cdot))$ is concave over $\Omega$.
\end{definition}
The following property of exp-concave functions  will be used later \citep[Lemma 3]{ML:Hazan:2007}.
\begin{lemma} \label{lem:exp} For a function $f:\Omega \mapsto \R$, where $\Omega$ has diameter $D$, such that $\forall \w \in \Omega$, $\|\nabla f(\w)\|\leq G$ and $\exp(-\alpha f(\cdot))$ is concave, the following holds for $\beta = \frac{1}{2} \min\{\frac{1}{4GD}, \alpha \}$:
\[
f(\y) \geq f(\x)+  \langle \nabla f(\x), \y -\x  \rangle + \frac{\beta}{2}   \langle \nabla f(\x), \y -\x  \rangle^2,  \  \forall \x, \y \in \Omega.
\]
\end{lemma}

\subsubsection{Technical challenge}
Before introducing the proposed algorithm, we discuss the technical challenge of minimizing the adaptive regret of multiple types of convex functions simultaneously. All the existing adaptive algorithms \citep{Adaptive:Hazan,Adaptive:ICML:15,Improved:Strongly:Adaptive} share the same framework and contain $3$ components:
\begin{compactitem}
  \item An expert-algorithm, which is able to minimize the static regret of a specific type of function;
  \item A set of intervals, each of which is associated with an expert-algorithm that minimizes the regret of that interval;
  \item A meta-algorithm, which combines the predictions of active experts in each round.
\end{compactitem}
To design a universal algorithm, a straightforward way is to use a universal method for static regret, such as MetaGrad \citep{NIPS2016_6268}, as the expert-algorithm. In this way, the expert-algorithm is able to handle the uncertainty of functions. However, the challenge lies in the design of the meta-algorithm, because the meta-algorithms used by previous studies also lack universality. For example, the meta-algorithm of  \citet{Adaptive:Hazan} is able to deliver a tight meta-regret for exp-concave functions, but a loose one for general convex functions. Similarly,  meta-algorithms of \citet{Adaptive:ICML:15} and \citet{Improved:Strongly:Adaptive} incur at least $\Theta(\sqrt{\tau})$ meta-regret for intervals of length $\tau$, which is tolerable for convex functions but suboptimal for exp-concave functions.

Instead of using MetaGrad as a black-box subroutine, we dig into this algorithm and modify it to minimize the adaptive regret directly. MetaGrad itself is a two-layer algorithm, which runs multiple expert-algorithms, each with a different learning rate, and combines them with a meta-algorithm named as Tilted Exponentially Weighted Average (TEWA). To address the aforementioned challenge, we extend TEWA to support sleeping experts so that it can minimize the adaptive regret.
The advantage of TEWA is that its meta-regret only depends on the number of experts instead of the length of the interval, e.g., Lemma 4 of \citet{NIPS2016_6268}, and thus does not affect the optimality of the regret. The extension of TEWA to sleeping experts is the main technical contribution of this paper.

\subsection{A parameter-free and adaptive algorithm for exp-concave functions}
Recall that our  goal is to design a universal algorithm for minimizing the adaptive regret of general convex,  exp-concave, and strongly convex functions simultaneously. However, to facilitate understanding, we  start with a simpler question: How to minimize the adaptive regret of exp-concave functions, without knowing the modulus of exp-concavity? By proposing a novel algorithm to answer the above question, we present the main techniques used in our paper. Then, we extend that algorithm to support other types of functions in the next section. Since our algorithm is built upon MetaGrad, we first review its key steps below.
\subsubsection{Review of MetaGrad}
The reason that MetaGrad can minimize the regret of $\alpha$-exp-concave functions without knowing the value of $\alpha$ is because it enjoys a second-order regret bound \citep{pmlr-v35-gaillard14}:
\begin{equation} \label{eqn:metagrad:1}
\sum_{t=1}^T f_t(\w_t) -  \sum_{t=1}^T f_t(\w) \leq \sum_{t=1}^T  \langle \nabla f_t(\w_t), \w_t -\w \rangle = O\left(\sqrt{V_T d \log T}  + d \log T \right)
\end{equation}
where $V_T=\sum_{t=1}^T \langle \nabla f_t(\w_t), \w_t -\w \rangle^2$. Besides, Lemma~\ref{lem:exp} implies
\begin{equation} \label{eqn:metagrad:2}
\sum_{t=1}^T f_t(\w_t) -  \sum_{t=1}^T f_t(\w) \leq \sum_{t=1}^T  \langle \nabla f_t(\w_t), \w_t -\w \rangle  - \frac{\beta}{2} \sum_{t=1}^T \langle \nabla f_t(\w_t), \w_t -\w \rangle^2.
\end{equation}
Combining (\ref{eqn:metagrad:1}) with (\ref{eqn:metagrad:2}) and applying the AM-GM inequality, we immediately obtain
\[
\sum_{t=1}^T f_t(\w_t) -  \sum_{t=1}^T f_t(\w) = O\left( \frac{d}{\beta}  \log T \right)=O\left( \frac{d}{\alpha}  \log T \right).
\]
From the above discussion, it becomes clear that if we can establish a second-order regret bound for any interval $[p, q]\subseteq [T]$, we are able to minimize the adaptive regret even when $\alpha$ is unknown.

The way that MetaGrad attains the regret bound in (\ref{eqn:metagrad:1}) is to run a set of experts, each of which minimizes a surrogate loss parameterized by a learning rate $\eta$
\begin{equation} \label{eqn:metagrad:3}
\ell_t^\eta(\w)= -\eta \langle \nabla f_t(\w_t),  \w_t -\w \rangle + \eta^2 \langle \nabla f_t(\w_t),  \w_t -\w \rangle^2
\end{equation}
and then combine the outputs of experts by a meta-algorithm named as Tilted Exponentially Weighted Average (TEWA). Specifically, it creates an expert $E^\eta$ for each $\eta$ in
\begin{equation} \label{eqn:metagrad:4}
\S(T)=\left\{ \frac{2^{-i}}{5 DG} \ \left |\ i=0,1,\ldots, \left\lceil \frac{1}{2} \log_2 T\right\rceil \right. \right\}
\end{equation}
and thus maintains $1+ \lceil \frac{1}{2} \log_2 T\rceil=O(\log T)$ experts during the learning process. By simultaneously considering multiple learning rates, MetaGrad is able to deal with the uncertainty of $V_T$.  Since the surrogate loss $\ell_t^\eta(\cdot)$ is exp-concave,  a variant of ONS is used as the expert-algorithm. Let $\w_t^\eta$ be the output of expert $E^\eta$ in the $t$-th round. MetaGrad calculates the final output $\w_t$ according to TEWA:
\begin{equation} \label{eqn:metagrad:5}
\w_t = \frac{\sum_\eta \pi_t^\eta \eta \w_t^\eta }{\sum_\eta \pi_t^\eta \eta}
\end{equation}
where $\pi_t^\eta \propto \exp(-\sum_{i=1}^{t-1} \ell_i^\eta(\w_i^\eta))$.
\subsubsection{Our approach}
In this section, we discuss how to minimize the adaptive regret by extending MetaGrad. Following the idea of sleeping experts \citep{Freund:1997:UCP}, the most straightforward way is to create $1+ \lceil \frac{1}{2} \log_2 (q-p+1)\rceil$ experts for each interval $[p,q] \in [T]$, and combine them with a meta-algorithm that supports sleeping experts. However, this simple approach is inefficient because the total number of experts is on the order of $O(T^2 \log T)$. To control the number of experts, we make use of the geometric covering (GC) intervals \citep{Adaptive:ICML:15} defined as
 \[
 \I= \bigcup_{k \in \N \cup \{0\}} \I_k,
 \]
where
  \[
  \I_k=\left\{ [ i \cdot 2^k, (i+1) \cdot 2^k -1]: i \in \N\right\}, \  k \in \N \cup \{0\}.
   \]
   A graphical illustration of GC intervals is given in Fig.~\ref{fig:interval:saol}. We observe that each $\I_k$ is a partition of $\N \setminus \{1,\cdots, 2^k-1 \}$ to consecutive intervals of length $2^k$.   Note that  GC intervals can be generated on the fly, so we do not need to fix the horizon $T$.
\begin{figure*}
\centering
\begin{tabular}{@{}c@{\hspace{0.9ex}}*{17}{@{\hspace{0.5ex}}c}@{\hspace{0.9ex}}c@{}}
$t$ & 1 & 2 & 3 & 4 & 5 & 6 & 7 & 8 & 9 & 10 &11 &12 & 13 & 14 &15 & 16 & 17  &$\cdots$ \\
 $\I_0$ & [\quad ] & [\quad ] &  [\quad  ] & [\quad ] & [\quad ] & [\quad ] & [\quad ] & [\quad ] & [\quad ] & [\quad ] & [\quad ] &[\quad ] &[\quad ] & [\quad ] & [\quad ] & [\quad ]& [\quad ]& $\cdots$   \\
 $\I_1$ &  & [\quad  \phantom{]}& \phantom{[}\quad ] & [\quad \phantom{]} & \phantom{[}\quad ] & [\quad \phantom{]} & \phantom{[}\quad ] & [\quad \phantom{]} & \phantom{[}\quad ] & [\quad \phantom{]} & \phantom{[}\quad ] &[\quad \phantom{]} &\phantom{[}\quad ] & [\quad \phantom{]} & \phantom{[}\quad ] & [\quad \phantom{]} & \phantom{[}\quad ]& $\cdots$   \\
 $\I_2$  &  & &  & [\quad \phantom{]} & \phantom{[}\quad \phantom{]} & \phantom{[}\quad \phantom{]} & \phantom{[}\quad ] & [\quad \phantom{]} & \phantom{[}\quad \phantom{]} & \phantom{[}\quad \phantom{]} & \phantom{[}\quad ] &[\quad \phantom{]} &\phantom{[}\quad \phantom{]} & \phantom{[}\quad \phantom{]} & \phantom{[}\quad ] & [\quad \phantom{]} & \phantom{[}\quad \phantom{]} &$\cdots$   \\
  $\I_3$   &  & &  &  &  &  &  & [\quad \phantom{]} & \phantom{[}\quad \phantom{]} & \phantom{[}\quad \phantom{]} & \phantom{[}\quad \phantom{]} &\phantom{[}\quad \phantom{]} &\phantom{[}\quad \phantom{]} & \phantom{[}\quad \phantom{]} & \phantom{[}\quad ] & [\quad \phantom{]} & \phantom{[}\quad \phantom{]}& $\cdots$   \\
$\I_4$   &  & &  &  &  &  &  &  &  &  &  & & &  &  & [\quad \phantom{]} & \phantom{[}\quad \phantom{]}& $\cdots$   \\
\end{tabular}
\caption{Geometric covering (GC) intervals of \citet{Adaptive:ICML:15}.}
\label{fig:interval:saol}
\end{figure*}

Then, we only focus on intervals in $\I$. For each interval $I=[r,s] \in \I$, we will create $1+ \lceil \frac{1}{2} \log_2 (s-r+1)\rceil$ experts, each of which minimizes one surrogate loss in $\{ \ell_t^\eta(\w)|  \eta \in \S(s-r+1)\}$ during $I$, where $\ell_t^\eta(\cdot)$ and $\S(\cdot)$ are defined in (\ref{eqn:metagrad:3}) and (\ref{eqn:metagrad:4}), respectively. These experts become active in round $r$ and will be removed forever after round $s$. 
Since the number of intervals that contain $t$ is $\lfloor \log_2 t \rfloor+1$ \citep{Adaptive:ICML:15},  the number of active experts is at most
\[
\left(\lfloor \log_2 t \rfloor+1\right) \left(1+ \left\lceil \frac{1}{2} \log_2 t \right \rceil\right)= O(\log^2 t).
\]
So, the number of active experts is larger than that of MetaGrad by a logarithmic factor, which is the price paid in computations for the adaptivity to every interval.

Finally, we need to specify how to combine the outputs of active experts. From the construction of the surrogate loss in (\ref{eqn:metagrad:3}), we observe that each expert uses its own loss, which is very different from the traditional setting of prediction with expert advice in which all experts use the same loss. As a result, existing meta-algorithms for adaptive regret \citep{Adaptive:ICML:15,Improved:Strongly:Adaptive}, which assume a fixed loss function, cannot be directly applied. In the literature, the setting that each expert uses a loss function that may be different from the loss functions used by the other experts has been studied by \citet{Expert_Evaluators}, who name it as prediction with expert evaluators' advice. Furthermore, they have proposed a general conversion to the sleeping expert case by only considering active rounds in the calculation of the cumulative losses.  Following this idea, we extend TEWA, the meta-algorithm of MetaGrad, to sleeping experts.

\begin{algorithm}[tb]
   \caption{A Parameter-free and Adaptive algorithm for Exp-concave functions (PAE)}
   \label{alg:PAE}
\begin{algorithmic}[1]
\STATE $\A_0=\emptyset$
   \FOR{$t=1$ {\bfseries to} $T$}
   \FORALL{$I \in \I$ that starts from $t$ }
    \FORALL{$\eta \in \S(|I|)$}
    \STATE Create an expert $E_I^\eta$ by running an instance of ONS to minimize $\ell_t^\eta(\cdot)$ during $I$, and set $L_{t-1,I}^\eta=0$
    \STATE Add $E_I^\eta$ to the set of active experts: $\A_t=\A_{t-1} \cup \{E_I^\eta\}$
   \ENDFOR
   \ENDFOR
   \STATE Receive output $\w_{t,J}^\eta$ from each expert $E_J^\eta \in \A_t$
   \STATE Submit $\w_t$ in (\ref{eqn:wt})
   \STATE Observe the loss $f_t(\cdot)$ and evaluate the gradient $\nabla f_t(\w_t)$
   \FORALL{$E_J^\eta \in \A_t$}
   \STATE Update $L_{t,J}^\eta=L_{t-1,J}^\eta + \ell_t^\eta(\w_{t,J}^\eta) $
   \STATE Pass the surrogate loss $\ell_t^\eta(\cdot)$ to expert $E_J^\eta$
   \ENDFOR
   \STATE Remove experts whose ending times are $t$ from $\A_t$
   \ENDFOR
\end{algorithmic}
\end{algorithm}

Our Parameter-free and Adaptive algorithm for Exp-concave functions (PAE) is summarized in Algorithm~\ref{alg:PAE}. In the  $t$-th round, we first create an expert  $E_I^\eta$ for each interval $I \in \I$ that starts from $t$ and each $\eta \in \S(|I|)$, where $\S(\cdot)$ is defined in (\ref{eqn:metagrad:4}), and introduce a variable $L_{t-1,I}^\eta$ to record the cumulative loss of  $E_I^\eta$ (Step 5). The expert $E_I^\eta$ is an instance of ONS \citep{ML:Hazan:2007} that minimizes $\ell_t^\eta(\cdot)$ during interval $I$. We also maintain a set $\A_t$ consisting of all the active experts (Step 6). Denote the prediction of expert $E_J^\eta$ at round $t$ as $\w_{t,J}^\eta$. In Step 9, PAE collects the predictions of all the active experts, and then submits the following solution in Step 10:
\begin{equation}\label{eqn:wt}
   \w_t = \frac{1}{\sum_{E_J^\eta \in \A_t} \exp(-L_{t-1,J}^\eta) \eta} \sum_{E_J^\eta \in \A_t}  \exp(-L_{t-1,J}^\eta) \eta \w_{t,J}^\eta .
\end{equation}
Compared with TEWA in (\ref{eqn:metagrad:5}), we observe that (\ref{eqn:wt}) focuses on active experts and ignores  inactive ones. Although the extension is inspired by \citet{Expert_Evaluators}, our analysis is different because the surrogate losses take a special form of (\ref{eqn:metagrad:3}). Specifically, we exploit the special structure of losses and apply a simple inequality \citep[Lemma 1]{Cesa-Bianchi2005}. In this way, we do not need to introduce advanced concepts such as the mixability  of \citet{Expert_Evaluators}. The analysis is still challenging because of the dynamic change of active experts. In Step 11, PAE observes the loss $f_t(\cdot)$ and evaluates the gradient $\nabla f_t(\w_t)$ to construct the surrogate loss. In Step 13, it updates the cumulative loss of each active expert, and in Step 14 passes the surrogate loss to each expert such that it can make predictions for the next round. In Step 16, PAE removes experts whose ending times are $t$ from $\A_t$.

 Next, we present the expert-algorithm.  It is easy to verify that the surrogate loss $\ell_t^\eta(\cdot)$ in (\ref{eqn:metagrad:3}) has the following property \citep[Lemma 2]{Adaptive:Maler}.
\begin{lemma} \label{lem:concave} Under Assumptions~\ref{ass:1} and \ref{ass:2}, $\ell_t^\eta(\cdot)$ in (\ref{eqn:metagrad:3}) is $1$-exp-concave, and
\[
\max_{\w \in \Omega}\|\nabla \ell_t^\eta(\w)\| \leq \frac{7}{25D}, \ \forall \eta \leq \frac{1}{5GD}.
\]
\end{lemma}
Thus, we can apply online Newton step (ONS) \citep{ML:Hazan:2007} as the expert-algorithm to minimize  $\ell_t^\eta(\cdot)$ during interval $I$. We provide the procedure of expert $E_{I}^\eta$ in Algorithm~\ref{alg:ONS}. The generalized projection $\Pi_\Omega^A(\cdot)$ associated with a positive semidefinite matrix $A $ is defined as
\[
\Pi_\Omega^A(\x)=\argmin_{\w \in \Omega} (\w-\x)^\top A (\w-\x)
\]
which is used in Step 6 of Algorithm~\ref{alg:ONS}.
\begin{algorithm}[tb]
   \caption{Expert $E_{I}^\eta$: Online Newton Step (ONS)}
   \label{alg:ONS}
\begin{algorithmic}[1]
   \STATE {\bfseries Input:} Interval $I=[r,s]$, $\eta$
   \STATE Let $\w_{r,I}^\eta$ be any point in $\Omega$
   \STATE $\beta=\frac{1}{2}\min\left( \frac{1}{4D \frac{7}{25D}}, 1 \right)=\frac{25}{56}$, $\Sigma_{r-1}=\frac{1}{\beta^2 D^2} I$
   \FOR{$t=r$ {\bfseries to} $s$}
   \STATE Update
\[
\Sigma_t=\Sigma_{t-1} + \nabla \ell_t^\eta(\w_{t,I}^\eta) \nabla \ell_t^\eta(\w_{t,I}^\eta)^\top
\]
where
\[
\nabla \ell_t^\eta(\w_{t,I}^\eta)= \eta \nabla f_t(\w_t) + 2 \eta^2 \left \langle \nabla f_t(\w_t), \w_{t,I}^\eta - \w_t \right \rangle \nabla f_t(\w_t)
\]
   \STATE Calculate
   \[
   \w_{t+1,I}^\eta = \Pi_\Omega^{\Sigma_t}\left( \w_{t,I}^\eta - \frac{1}{\beta} \Sigma_t^{-1} \nabla \ell_t^\eta(\w_{t,I}^\eta) \right)
   \]
   \ENDFOR
\end{algorithmic}
\end{algorithm}

We present the theoretical guarantee of PAE below.
\begin{thm}\label{thm:PAE} Under Assumptions~\ref{ass:1} and \ref{ass:2}, for any interval $[p,q] \subseteq [T]$ and any $\w \in \Omega$, PAE satisfies
\[
\begin{split}
 \sum_{t=p}^q\langle \nabla f_t(\w_t),  \w_t -\w \rangle \leq  10 DG  a(p,q) b(p,q) + 3 \sqrt{a(p,q) b(p,q)} \sqrt{\sum_{t=p}^q\langle \nabla f_t(\w_t),  \w_t -\w \rangle^2}
\end{split}
\]
where
\begin{align}
a(p,q)=&2\log_2(2q) + 5 d \log(q-p+2)+5, \label{eqn:a:fun}\\
b(p,q)=&2\lceil\log_2(q-p+2)\rceil . \label{eqn:b:fun}
\end{align}
Furthermore, if all the online functions are $\alpha$-exp-concave, we have
\[
\sum_{t=p}^q f_t(\w_t) - \sum_{t=p}^q f_t(\w)  \leq \left(10 DG    + \frac{9}{2 \beta} \right) a(p,q) b(p,q)= O\left( \frac{d \log q \log(q-p)}{\alpha} \right)
\]
where $\beta = \frac{1}{2} \min\{\frac{1}{4GD}, \alpha \}$.
\end{thm}
\paragraph{Remark} Theorem~\ref{thm:PAE} indicates that PAE enjoys a second-order regret bound for any interval, which in turn implies a small regret for exp-concave functions. Specifically, for $\alpha$-exp-concave functions, PAE satisfies $\SAReg(T,\tau)= O(\frac{d}{\alpha}\log \tau \log T )$, which matches the regret of efficient FLH \citep{Adaptive:Hazan}. This is a remarkable result given the fact that PAE is \emph{agnostic} to $\alpha$.
\begin{algorithm}[tb]
   \caption{A Universal algorithm for Minimizing the Adaptive regret (UMA)}
   \label{alg:UMA}
\begin{algorithmic}[1]
\STATE $\A_0=\widehat{\A}_0=\emptyset$
   \FOR{$t=1$ {\bfseries to} $T$}
   \FORALL{$I \in \I$ that starts from $t$ }
    \FORALL{$\eta \in \S(|I|)$}
    \STATE Create an expert $E_I^\eta$ by running an instance of ONS to minimize $\ell_t^\eta(\cdot)$ during $I$, and set $L_{t-1,I}^\eta=0$
     \STATE Add $E_I^\eta$ to the set of active experts: $\A_t=\A_{t-1} \cup \{E_I^\eta\}$
     \STATE Create an expert $\widehat{E}_I^\eta$ by running an instance of AOGD to minimize $\hat{\ell}_t^\eta(\cdot)$ during $I$, and set $\widehat{L}_{t-1,I}^\eta=0$
     \STATE Add $\widehat{E}_I^\eta$ to the set of active experts: $\widehat{\A}_t=\widehat{\A}_{t-1} \cup \{\widehat{E}_I^\eta\}$
   \ENDFOR
   \ENDFOR
   \STATE Receive output $\w_{t,J}^\eta$ from each expert $E_J^\eta \in \A_t$ and $\wh_{t,J}^\eta$ from each  expert $\widehat{E}_J^\eta \in \widehat{\A}_t$
   \STATE Submit $\w_t$ in (\ref{eqn:wt:new})
   \STATE Observe the loss $f_t(\cdot)$ and evaluate the gradient $\nabla f_t(\w_t)$
   \FORALL{$E_J^\eta \in \A_t$}
   \STATE Update $L_{t,J}^\eta=L_{t-1,J}^\eta + \ell_t^\eta(\w_{t,J}^\eta) $
   \STATE Pass the surrogate loss $\ell_t^\eta(\cdot)$ to expert $E_J^\eta$
   \ENDFOR
   \FORALL{$\widehat{E}_J^\eta \in \widehat{\A}_t$}
   \STATE Update $\widehat{L}_{t,J}^\eta=\widehat{L}_{t-1,J}^\eta + \hat{\ell}_t^\eta(\w_{t,J}^\eta) $
   \STATE Pass the surrogate loss $\hat{\ell}_t^\eta(\cdot)$ to expert $\widehat{E}_J^\eta$
   \ENDFOR
   \STATE Remove experts whose ending times are $t$ from $\A_t$ and $\widehat{\A}_t$
   \ENDFOR
\end{algorithmic}
\end{algorithm}

\begin{algorithm}[t]
   \caption{Expert $\widehat{E}_{I}^\eta$: Adaptive Online Gradient Descent (AOGD)}
   \label{alg:AOGD}
\begin{algorithmic}[1]
   \STATE {\bfseries Input:} Interval $I=[r,s]$, $\eta$
   \STATE Let $\wh_{r,I}^\eta$ be any point in $\Omega$
   \FOR{$t=r$ {\bfseries to} $s$}
   \STATE Update
   \[
   \wh_{t+1,I}^\eta = \Pi_\Omega\left( \wh_{t,I}^\eta - \frac{1}{\alpha_t} \nabla \hat{\ell}_t^\eta(\wh_{t,I}^\eta) \right)
   \]
   where
\[
\begin{split}
\nabla \hat{\ell}_t^\eta(\wh_{t,I}^\eta)= &\eta \nabla f_t(\w_t) + 2 \eta^2 \|\nabla f_t(\w_t)\|^2  ( \wh_{t,I}^\eta - \w_t)\\
\alpha_t=& 2 \eta^2G^2 + 2 \eta^2 \sum_{i=r}^t  \|\nabla f_i(\w_i)\|^2
\end{split}
\]
   \ENDFOR
\end{algorithmic}
\end{algorithm}

\subsection{A universal algorithm for minimizing the adaptive regret}
In this section, we extend PAE to support strongly convex functions and general convex functions. Inspired by \citet{AAAI:2020:Wang}, we introduce a new surrogate loss to handle strong convexity:
\begin{equation} \label{eqn:ell:hat}
\hat{\ell}_t^\eta(\w)= -\eta \langle \nabla f_t(\w_t),  \w_t -\w \rangle + \eta^2 \|\nabla f_t(\w_t)\|^2 \|\w_t -\w \|^2
\end{equation}
which is also parameterized by $\eta>0$.  Our goal is to attain another second-order type of regret bound
\begin{equation} \label{eqn:second:new}
 \sum_{t=p}^q f_t(\w_t) - \sum_{t=p}^q f_t(\w)\leq  \sum_{t=p}^q\langle \nabla f_t(\w_t),  \w_t -\w \rangle =\widetilde{O}\left(  \sqrt{\sum_{t=p}^q \|  \w_t -\w \|^2} \right)
\end{equation}
for any interval $[p,q] \subseteq T$. Combining (\ref{eqn:second:new}) with Definition~\ref{def:strong}, we can establish a tight regret bound for $\lambda$-strongly convex functions over any interval without knowing the value of $\lambda$. Furthermore, upper bounding $\sum_{t=p}^q \|  \w_t -\w \|^2$ in (\ref{eqn:second:new}) by $(q-p+1)D^2$, we obtain a  regret bound for general convex functions over any interval.  As a result, there is no need to add additional surrogate losses for general convex functions.

Our Universal algorithm for Minimizing the Adaptive regret (UMA) is summarized in Algorithm~\ref{alg:UMA}. UMA is a natural extension of PAE by incorporating the new surrogate loss $\hat{\ell}_t^\eta(\cdot)$. The overall procedure of UMA is very similar to PAE, except that the number of experts doubles and the weighting formula is modified accordingly. Specifically, in each round $t$, we further create an expert $\widehat{E}_I^\eta$ for each interval $I \in \I$ that starts from $t$ and each $\eta \in \S(|I|)$. $\widehat{E}_I^\eta$  is an instance of AOGD  that is able to minimize $\hat{\ell}_t^\eta$ during interval $I$. We use $\widehat{L}_{t-1,I}^\eta$ to  represent the cumulative loss of $\widehat{E}_I^\eta$ till round $t-1$, and $\widehat{\A}_t$ to store all the active $\widehat{E}_I^\eta$'s.  Denote the prediction of expert $\widehat{E}_J^\eta$ at round $t$ as $\wh_{t,J}^\eta$. In Step 11, UMA receives predictions from experts in $\A_t$ and $\widehat{\A}_t$, and submits the following solution in Step 12:
\begin{equation}\label{eqn:wt:new}
   \w_t = \frac{\sum_{E_J^\eta \in \A_t} \exp(-L_{t-1,J}^\eta) \eta \w_{t,J}^\eta +  \sum_{\widehat{E}_J^\eta \in \widehat{\A}_t}\exp(-\widehat{L}_{t-1,J}^\eta) \eta \wh_{t,J}^\eta}{\sum_{E_J^\eta \in \A_t} \exp(-L_{t-1,J}^\eta) \eta +\sum_{\widehat{E}_J^\eta \in \widehat{\A}_t}\exp(-\widehat{L}_{t-1,J}^\eta) \eta}
\end{equation}
which is an extension of (\ref{eqn:wt}) to accommodate more experts.

Next, we present the expert-algorithm for the new surrogate loss $\hat{\ell}_t^\eta(\cdot)$ in (\ref{eqn:ell:hat}). It is easy to verify that the  $\hat{\ell}_t^\eta(\cdot)$ enjoys the following property  \citep[Lemma 3 and Lemma 4]{AAAI:2020:Wang}.
\begin{lemma} \label{lem:strong:convex} Under Assumptions~\ref{ass:1} and \ref{ass:2}, $\hat{\ell}_t^\eta(\cdot)$ in (\ref{eqn:ell:hat}) is $2\eta^2 \|\nabla f_t(\w_t)\|^2$-strongly convex, and
\[
\max_{\w \in \Omega}\|\nabla \hat{\ell}_t^\eta(\w)\| \leq  2\eta^2 \|\nabla f_t(\w_t)\|^2,   \ \forall \eta \leq \frac{1}{5GD}.
\]
\end{lemma}
Thus, although $\hat{\ell}_t^\eta(\cdot)$ is strongly convex, the modulus of strong convexity, i.e., $2 \eta^2 \|\nabla f_t(\w_t)\|^2 $ is not fixed. So, we choose AOGD \citep{NIPS2007_3319} instead of OGD  \citep{ML:Hazan:2007} as the expert-algorithm to minimize  $\hat{\ell}_t^\eta(\cdot)$ during interval $I$. We provide the procedure of expert $\widehat{E}_{I}^\eta$ in Algorithm~\ref{alg:AOGD}. The projection operator $\Pi_\Omega(\cdot)$ is defined as
\[
\Pi_\Omega(\x)=\argmin_{\w \in \Omega} \|\w-\x\| .
\]

Our analysis shows that UMA inherits the theoretical guarantee of PAE, and meanwhile is able to minimize the adaptive regret of general convex and strongly convex functions.

\begin{thm} \label{thm:main}
Under Assumptions~\ref{ass:1} and \ref{ass:2}, for any interval $[p,q] \subseteq [T]$ and any $\w \in \Omega$, UMA enjoys the theoretical guarantee of PAE in Theorem~\ref{thm:PAE}. Besides, it also satisfies
\begin{align}
 \sum_{t=p}^q\langle \nabla f_t(\w_t),  \w_t -\w \rangle \leq  &10 DG  \hat{a}(p,q) b(p,q) + 3 G\sqrt{\hat{a}(p,q) b(p,q)} \sqrt{\sum_{t=p}^q \| \w_t -\w \|^2}, \label{eqn:UMA:inequality:1}\\
\sum_{t=p}^q\langle \nabla f_t(\w_t),  \w_t -\w \rangle \leq  &10 DG  \hat{a}(p,q) b(p,q) + 21 DG\sqrt{\hat{a}(p,q)(q-p+1)} \label{eqn:UMA:inequality:2}
\end{align}
where $b(\cdot,\cdot)$ is given in (\ref{eqn:b:fun}), and
\begin{equation} \label{eqn:a:hat}
\hat{a}(p,q)= 1+ 2\log_2(2q) + \log (q-p+2) .
\end{equation}
Furthermore, if all the online functions are $\lambda$-strongly convex, we have
\[
\sum_{t=p}^q f_t(\w_t) - \sum_{t=p}^q f_t(\w)  \leq \left(10 DG    + \frac{9G^2}{2\lambda} \right) \hat{a}(p,q) b(p,q)= O\left( \frac{\log q \log(q-p)}{\lambda} \right).
\]
\end{thm}
\paragraph{Remark} First, (\ref{eqn:UMA:inequality:1}) shows that UMA is equipped with another second-order regret bound for any interval, leading to a small regret for strongly convex functions. Specifically, for $\lambda$-strongly convex functions, UMA achieves $\SAReg(T,\tau)= O(\frac{1}{\lambda} \log \tau \log T )$, which matches the regret of the efficient algorithm of \citet{Dynamic:Regret:Adaptive}. Second, (\ref{eqn:UMA:inequality:2}) manifests that UMA attains an $O(\sqrt{ \tau \log T})$ strongly adaptive regret for general convex functions, which again matches the state-of-the-art result of \citet{Improved:Strongly:Adaptive} exactly. Finally, because of the dual adaptivity, UMA can handle the tough case that the type of functions switches or the parameter of functions changes.

\section{Analysis}
In this section, we present proofs of main theorems.
\subsection{Proof of Theorem~\ref{thm:PAE}}
We start with the meta-regret of PAE over any interval in $\I$.
\begin{lemma}\label{lem:meta} Under Assumptions~\ref{ass:1} and \ref{ass:2}, for any interval $I=[r,s] \in \I$ and any $\eta \in \S(s-r+1)$, the meta-regret of PAE with respect to $E_I^\eta$ satisfies
\[
\sum_{t=r}^s \ell_t^\eta(\w_t) - \sum_{t=r}^s \ell_t^\eta(\w_{t,I}^\eta) =  - \sum_{t=r}^s \ell_t^\eta(\w_{t,I}^\eta) \leq 2\log_2(2s).
\]
\end{lemma}

Then, combining Lemma~\ref{lem:meta} with the regret of expert $E_I^\eta$, which is just the regret bound of ONS over $I$, we establish a second-order regret of PAE over any interval in $\I$.
\begin{lemma}\label{lem:second} Under Assumptions~\ref{ass:1} and \ref{ass:2}, for any interval $I=[r,s] \in \I$ and any $\w \in \Omega$, PAE satisfies
\begin{equation}\label{eqn:pae:sec:int}
\sum_{t=r}^s\langle \nabla f_t(\w_t),  \w_t -\w \rangle \leq   3 \sqrt{a(r,s)\sum_{t=r}^s \langle \nabla f_t(\w_t),  \w_t -\w \rangle^2 }+ 10 DG a(r,s)
\end{equation}
where $a(\cdot,\cdot)$ is defined in (\ref{eqn:a:fun}).
\end{lemma}

Based on the following property of GC intervals \citep[Lemma 1.2]{Adaptive:ICML:15}, we extend Lemma~\ref{lem:second} to any interval $[p,q] \subseteq [T]$.
\begin{lemma}\label{lem:GC:intervals} For any interval $[p,q] \subseteq [T]$, it can be partitioned into two sequences of disjoint and consecutive intervals, denoted by $I_{-m},\ldots,I_0 \in \I$ and $I_1,\ldots,I_n \in \I$, such that
\[
|I_{-i}|/ |I_{-i+1}| \leq 1/2, \ \forall i \geq 1
\]
and
\[
|I_i|/|I_{i-1}| \leq 1/2, \ \forall i \geq 2.
\]
\end{lemma}
From the above lemma, we conclude that $n \leq \lceil\log_2(q-p+2)\rceil$ because otherwise
\[
|I_1|+\cdots+|I_n| \geq 1+2+\ldots + 2^{n-1}= 2^n -1 > q-p+1 = |I|.
\]
Similarly, we have $m+1 \leq \lceil\log_2(q-p+2)\rceil$.

For any interval $[p,q] \subseteq [T]$, let  $I_{-m},\ldots,I_0 \in \I$ and $I_1,\ldots,I_n \in \I$ be the partition described in Lemma~\ref{lem:GC:intervals}. Then, we have
\begin{equation}\label{eqn:decom:regret}
\sum_{t=p}^q\langle \nabla f_t(\w_t),  \w_t -\w \rangle = \sum_{i=-m}^n \sum_{t\in I_i}\langle \nabla f_t(\w_t),  \w_t -\w \rangle.
\end{equation}
Combining with Lemma~\ref{lem:second}, we have
\begin{equation} \label{eqn:thm1:tmp}
\begin{split}
&\sum_{t=p}^q\langle \nabla f_t(\w_t),  \w_t -\w \rangle \\
\leq &\sum_{i=-m}^n  \left(3 \sqrt{a(p,q)\sum_{t\in I_i} \langle \nabla f_t(\w_t),  \w_t -\w \rangle^2 }+ 10 DG a(p,q) \right) \\
= & 10 DG (m+1+n) a(p,q) + 3 \sqrt{a(p,q)}  \sum_{i=-m}^n \sqrt{\sum_{t\in I_i} \langle \nabla f_t(\w_t),  \w_t -\w \rangle^2 } \\
\leq & 10 DG (m+1+n) a(p,q) + 3 \sqrt{(m+1+n)a(p,q)} \sqrt{\sum_{i=-m}^n \sum_{t\in I_i} \langle \nabla f_t(\w_t),  \w_t -\w \rangle^2}\\
=& 10 DG (m+1+n) a(p,q) + 3 \sqrt{(m+1+n)a(p,q)} \sqrt{\sum_{t=p}^q\langle \nabla f_t(\w_t),  \w_t -\w \rangle^2}\\
\leq & 10 DG  a(p,q) b(p,q) + 3 \sqrt{a(p,q) b(p,q)} \sqrt{\sum_{t=p}^q\langle \nabla f_t(\w_t),  \w_t -\w \rangle^2}.
\end{split}
\end{equation}

When all the online functions are $\alpha$-exp-concave, Lemma~\ref{lem:exp} implies
\[
\begin{split}
&\sum_{t=p}^q f_t(\w_t) -  \sum_{t=p}^q f_t(\w) \\
\leq & \sum_{t=p}^q  \langle \nabla f_t(\w_t), \w_t -\w \rangle  - \frac{\beta}{2} \sum_{t=p}^q \langle \nabla f_t(\w_t), \w_t -\w \rangle^2\\
\overset{\text{(\ref{eqn:thm1:tmp})}}{\leq} & 10 DG  a(p,q) b(p,q) + 3 \sqrt{a(p,q) b(p,q)} \sqrt{\sum_{t=p}^q\langle \nabla f_t(\w_t),  \w_t -\w \rangle^2}\\
& -\frac{\beta}{2} \sum_{t=p}^q \langle \nabla f_t(\w_t), \w_t -\w \rangle^2\\
\leq & \left(10 DG    + \frac{9}{2 \beta} \right) a(p,q) b(p,q).
\end{split}
\]

\subsection{Proof of Lemma~\ref{lem:meta}}
This lemma is an extension of Lemma 4 of \citet{NIPS2016_6268} to sleeping experts.  We first introduce the following inequality \citep[Lemma 1]{Cesa-Bianchi2005}.
\begin{lemma} \label{lem:inequality} For all $z\geq -\frac{1}{2}$, $\ln(1+z)\geq z-z^2$.
\end{lemma}
For any $\w \in \Omega$ and any $\eta \leq \frac{1}{5GD}$, we have
\[
\eta \langle \nabla f_t(\w_t),  \w_t -\w \rangle \geq -\eta \|\nabla f_t(\w_t)\|\|\w_t -\w \| \overset{\text{(\ref{eqn:domain}),(\ref{eqn:gradient})}}{\geq} -\frac{1}{5}.
\]
Then, according to Lemma~\ref{lem:inequality}, we have
\begin{equation}\label{eqn:lem:meta:1}
\begin{split}
\exp\left(-\ell_t^\eta(\w) \right) = &\exp\left(\eta \langle \nabla f_t(\w_t),  \w_t -\w \rangle - \eta^2 \langle \nabla f_t(\w_t),  \w_t -\w \rangle^2  \right) \\
\leq & 1+ \eta \langle \nabla f_t(\w_t),  \w_t -\w \rangle.
\end{split}
\end{equation}

Recall that $\A_t$ is the set of active experts in round $t$, and $L_{t,J}^\eta$ is the cumulative loss of expert $E_{J}^\eta$. We have

\begin{equation}\label{eqn:lem:meta:2}
\begin{split}
& \sum_{E_J^\eta \in \A_t} \exp(-L_{t,J}^\eta) = \sum_{E_J^\eta \in \A_t} \exp(-L_{t-1,J}^\eta) \exp\left(-\ell_t^\eta(\w_{t,J}^\eta) \right) \\
\overset{\text{(\ref{eqn:lem:meta:1})}}{\leq} & \sum_{E_J^\eta \in \A_t} \exp(-L_{t-1,J}^\eta)\left(1+ \eta \langle \nabla f_t(\w_t),  \w_t -\w_{t,J}^\eta \rangle \right)\\
 =& \sum_{E_J^\eta \in \A_t} \exp(-L_{t-1,J}^\eta) + \left \langle \nabla f_t(\w_t),  \sum_{E_J^\eta \in \A_t} \exp(-L_{t-1,J}^\eta) \eta \w_t -\sum_{E_J^\eta \in \A_t} \exp(-L_{t-1,J}^\eta) \eta \w_{t,J}^\eta \right \rangle\\
\overset{\text{(\ref{eqn:wt})}}{=} & \sum_{E_J^\eta \in \A_t} \exp(-L_{t-1,J}^\eta) .
\end{split}
\end{equation}
Summing (\ref{eqn:lem:meta:2}) over $t=1,\ldots,s$, we have
\[
\sum_{t=1}^s \sum_{E_J^\eta \in \A_t} \exp(-L_{t,J}^\eta)  \leq \sum_{t=1}^s  \sum_{E_J^\eta \in \A_t} \exp(-L_{t-1,J}^\eta)
\]
which can be rewritten as
\[
\begin{split}
&\sum_{E_J^\eta \in \A_s} \exp(-L_{s,J}^\eta) +\sum_{t=1}^{s-1} \left(\sum_{E_J^\eta \in \A_{t} \setminus \A_{t+1}} \exp(-L_{t,J}^\eta) +  \sum_{E_J^\eta \in \A_{t} \cap \A_{t+1}} \exp(-L_{t,J}^\eta) \right)\\
\leq &\sum_{E_J^\eta \in \A_1} \exp(-L_{0,J}^\eta)+\sum_{t=2}^s \left( \sum_{E_J^\eta \in \A_t \setminus \A_{t-1}} \exp(-L_{t-1,J}^\eta) + \sum_{E_J^\eta \in \A_t \cap \A_{t-1}} \exp(-L_{t-1,J}^\eta)\right)
\end{split}
\]
implying
\begin{equation}\label{eqn:lem:meta:3}
\begin{split}
&\sum_{E_J^\eta \in \A_s} \exp(-L_{s,J}^\eta) + \sum_{t=1}^{s-1}  \sum_{E_J^\eta \in \A_{t} \setminus \A_{t+1} } \exp(-L_{t,J}^\eta) \\
\leq  &\sum_{E_J^\eta \in \A_1} \exp(-L_{0,J}^\eta) + \sum_{t=2}^{s} \sum_{E_J^\eta \in \A_t \setminus \A_{t-1}} \exp(-L_{t-1,J}^\eta) \\
=& \sum_{E_J^\eta \in \A_1} \exp(0) + \sum_{t=2}^{s} \sum_{E_J^\eta \in \A_t \setminus \A_{t-1}} \exp(0)\\
=& |\A_1| + \sum_{t=2}^{s}| \A_t \setminus \A_{t-1}|.
\end{split}
\end{equation}

Note that $|\A_1| + \sum_{t=2}^{s}| \A_t \setminus \A_{t-1}|$ is the total number of experts created until round $s$. From the structure of GC intervals and (\ref{eqn:metagrad:4}), we have
\begin{equation}\label{eqn:lem:meta:4}
|\A_1| + \sum_{t=2}^{s}| \A_t \setminus \A_{t-1}| \leq s \left(\lfloor \log_2 s \rfloor+1\right) \left(1+ \left\lceil \frac{1}{2} \log_2 s \right \rceil\right) \leq 4s^2.
\end{equation}
From (\ref{eqn:lem:meta:3}) and (\ref{eqn:lem:meta:4}), we have
\[
\sum_{E_J^\eta \in \A_s} \exp(-L_{s,J}^\eta) + \sum_{t=1}^{s-1}  \sum_{E_J^\eta \in \A_{t} \setminus \A_{t+1} } \exp(-L_{t,J}^\eta) \leq 4s^2.
\]

Thus, for any interval $I=[r,s] \in \I$, we have
\[
\exp(-L_{s,I}^\eta) = \exp\left( - \sum_{t=r}^s \ell_t^\eta(\w_{t,I}^\eta) \right)\leq 4s^2
\]
which completes the proof.
\subsection{Proof of Lemma~\ref{lem:second}}
The analysis is similar to the proofs of Theorem 7 of \citet{NIPS2016_6268} and Theorem 1 of \citet{Adaptive:Maler}.

From Lemma~\ref{lem:concave} and Theorem 2 of \citet{ML:Hazan:2007}, we have the following expert-regret of $E_I^\eta$ \citep[Lemma 2]{Adaptive:Maler}.
\begin{lemma}\label{lem:expert} Under Assumptions~\ref{ass:1} and \ref{ass:2}, for any interval $I=[r,s] \in \I$ and any $\eta \in \S(s-r+1)$, the expert-regret of $E_I^\eta$ satisfies
\[
\sum_{t=r}^s \ell_t^\eta(\w_{t,I}^\eta) - \sum_{t=r}^s \ell_t^\eta(\w)  \leq 5 d \log(s-r+2)+5, \ \forall \w \in \Omega.
\]
\end{lemma}
Combining the regret bounds in Lemmas~\ref{lem:meta} and \ref{lem:expert}, we have
\[
\begin{split}
- \sum_{t=r}^s \ell_t^\eta(\w) = & \eta \sum_{t=r}^s \langle \nabla f_t(\w_t),  \w_t -\w \rangle - \eta^2 \sum_{t=r}^s \langle \nabla f_t(\w_t),  \w_t -\w \rangle^2 \\
\leq & 2\log_2(2s) + 5 d \log(s-r+2) +5
\end{split}
\]
for any $\eta \in \S(s-r+1)$. Thus,
\begin{equation} \label{eqn:lem:second:1}
\sum_{t=r}^s\langle \nabla f_t(\w_t),  \w_t -\w \rangle \leq \frac{ 2\log_2(2s) + 5 d \log(s-r+2)+5}{\eta} + \eta \sum_{t=r}^s \langle \nabla f_t(\w_t),  \w_t -\w \rangle^2
\end{equation}
for any $\eta \in \S(s-r+1)$.

Let $a(r,s)=2\log_2(2s) + 5 d \log(s-r+2)+5 \geq 2$. Note that the optimal $\eta_*$ that minimizes the R.H.S.~of (\ref{eqn:lem:second:1}) is
\[
\eta_*=\sqrt{\frac{a(r,s)}{\sum_{t=r}^s \langle \nabla f_t(\w_t),  \w_t -\w \rangle^2}} \geq \frac{\sqrt{2}}{GD \sqrt{s-r+1}}.
\]
Recall that
\[
\S(s-r+1)= \left\{ \frac{2^{-i}}{5 DG} \ \left |\ i=0,1,\ldots, \left\lceil \frac{1}{2} \log_2 (s-r+1)\right\rceil \right. \right\}.
\]
If $\eta_* \leq \frac{1}{5DG}$, there must exist an $\eta \in \S(s-r+1)$ such that
\[
\eta \leq \eta_* \leq 2\eta.
\]
Then, (\ref{eqn:lem:second:1}) implies
\begin{equation} \label{eqn:lem:second:2}
\begin{split}
\sum_{t=r}^s\langle \nabla f_t(\w_t),  \w_t -\w \rangle \leq &  2\frac{ a(r,s)}{\eta_*} + \eta_* \sum_{t=r}^s \langle \nabla f_t(\w_t),  \w_t -\w \rangle^2\\
 =& 3 \sqrt{a(r,s)\sum_{t=r}^s \langle \nabla f_t(\w_t),  \w_t -\w \rangle^2 }.
\end{split}
\end{equation}
On the other hand, if $\eta_* \geq \frac{1}{5DG}$,  we have
\[
\sum_{t=r}^s \langle \nabla f_t(\w_t),  \w_t -\w \rangle^2 \leq  25 D^2 G^2 a(r,s).
\]
Then, (\ref{eqn:lem:second:1}) with $\eta=\frac{1}{5DG}$ implies
\begin{equation} \label{eqn:lem:second:3}
\sum_{t=r}^s\langle \nabla f_t(\w_t),  \w_t -\w \rangle \leq  5DG a(r,s)+ 5 DG a(r,s)   = 10 DG a(r,s).
\end{equation}

We complete the proof by combining (\ref{eqn:lem:second:2}) and (\ref{eqn:lem:second:3}).

\subsection{Proof of Theorem~\ref{thm:main}}
We first show the meta-regret of UMA, which is similar to Lemma~\ref{lem:meta} of PAE.
\begin{lemma}\label{lem:meta:UMA} Under Assumptions~\ref{ass:1} and \ref{ass:2}, for any interval $I=[r,s] \in \I$ and any $\eta \in \S(s-r+1)$, the meta-regret of UMA satisfies
\[
\begin{split}
\sum_{t=r}^s \ell_t^\eta(\w_t) - \sum_{t=r}^s \ell_t^\eta(\w_{t,I}^\eta) =  - \sum_{t=r}^s \ell_t^\eta(\w_{t,I}^\eta) \leq 2\log_2(2s),\\
\sum_{t=r}^s \hat{\ell}_t^\eta(\w_t) - \sum_{t=r}^s \hat{\ell}_t^\eta(\wh_{t,I}^\eta) =  - \sum_{t=r}^s \hat{\ell}_t^\eta(\wh_{t,I}^\eta) \leq 2\log_2(2s).
\end{split}
\]
\end{lemma}

Then, combining with the expert-regret of $E_I^\eta$ and $\widehat{E}_I^\eta$, we prove the following second-order regret of UMA over any interval in $\I$, which is similar to Lemma~\ref{lem:second} of PAE.
\begin{lemma}\label{lem:second:UMA} Under Assumptions~\ref{ass:1} and \ref{ass:2}, for any interval $I=[r,s] \in \I$ and any $\w \in \Omega$, UMA satisfies
\begin{align}
\sum_{t=r}^s\langle \nabla f_t(\w_t),  \w_t -\w \rangle \leq  & 3 \sqrt{a(r,s)\sum_{t=r}^s \langle \nabla f_t(\w_t),  \w_t -\w \rangle^2 }+ 10 DG a(r,s),  \label{eqn:second:1} \\
\sum_{t=r}^s\langle \nabla f_t(\w_t),  \w_t -\w \rangle \leq  & 3G \sqrt{\hat{a}(r,s)\sum_{t=r}^s  \|\w_t -\w\|^2 }+ 10 DG \hat{a}(r,s)  \label{eqn:second:2}
\end{align}
where $a(\cdot,\cdot)$  and $\hat{a}(\cdot,\cdot)$ are defined in (\ref{eqn:a:fun}) and  (\ref{eqn:a:hat}), respectively.
\end{lemma}

Based on the property of GC intervals \citep[Lemma 1.2]{Adaptive:ICML:15}, we extend Lemma~\ref{lem:second:UMA} to any interval $[p,q] \subseteq [T]$. Notice that (\ref{eqn:second:1}) is the same as (\ref{eqn:pae:sec:int}), so  Theorem~\ref{thm:PAE} also holds for UMA. In the following, we prove (\ref{eqn:UMA:inequality:1}) in a similar way. Combining (\ref{eqn:decom:regret}) with (\ref{eqn:second:2}), we have
\begin{equation} \label{eqn:UMA:inequality:3}
\begin{split}
&\sum_{t=p}^q\langle \nabla f_t(\w_t),  \w_t -\w \rangle \\
\leq &\sum_{i=-m}^n  \left(3 G\sqrt{\hat{a}(p,q)\sum_{t\in I_i} \|\w_t -\w \|^2 }+ 10 DG \hat{a}(p,q) \right) \\
= & 10 DG (m+1+n) \hat{a}(p,q) + 3 G\sqrt{\hat{a}(p,q)}  \sum_{i=-m}^n \sqrt{\sum_{t\in I_i} \|\w_t -\w \|^2 } \\
\leq & 10 DG (m+1+n) \hat{a}(p,q) + 3G \sqrt{(m+1+n)\hat{a}(p,q)} \sqrt{\sum_{i=-m}^n \sum_{t\in I_i} \|\w_t -\w \|^2}\\
=& 10 DG (m+1+n) \hat{a}(p,q) + 3G \sqrt{(m+1+n)\hat{a}(p,q)} \sqrt{\sum_{t=p}^q\|\w_t -\w \|^2}\\
\leq & 10 DG  \hat{a}(p,q) b(p,q) + 3G \sqrt{\hat{a}(p,q) b(p,q)} \sqrt{\sum_{t=p}^q\|\w_t -\w \|^2}.
\end{split}
\end{equation}

We proceed to prove (\ref{eqn:UMA:inequality:2}). If we upper bound $\sum_{t=p}^q \| \w_t -\w \|^2$ in  (\ref{eqn:UMA:inequality:1})  by $D^2 (q-p+1)$, we arrive at
\[
\sum_{t=p}^q\langle \nabla f_t(\w_t),  \w_t -\w \rangle \leq  10 DG  \hat{a}(p,q) b(p,q) + 3 DG\sqrt{\hat{a}(p,q) b(p,q)} \sqrt{q-p+1}
\]
which is worse than (\ref{eqn:UMA:inequality:2}) by a $\sqrt{b(p,q)}$ factor. To avoid this factor,  we use a different way to simplify (\ref{eqn:UMA:inequality:3}):
\begin{equation} \label{eqn:UMA:inequality:4}
\begin{split}
&\sum_{t=p}^q\langle \nabla f_t(\w_t),  \w_t -\w \rangle \\
\leq &\sum_{i=-m}^n  \left(3 G\sqrt{\hat{a}(p,q)\sum_{t\in I_i} \|\w_t -\w \|^2 }+ 10 DG \hat{a}(p,q) \right) \\
= & 10 DG (m+1+n) \hat{a}(p,q) + 3 G\sqrt{\hat{a}(p,q)}  \sum_{i=-m}^n \sqrt{\sum_{t\in I_i} \|\w_t -\w \|^2 } \\
\leq & 10 \hat{a}(p,q) b(p,q)+ 3 DG\sqrt{\hat{a}(p,q)}   \sum_{i=-m}^n \sqrt{|I_i|}.
\end{split}
\end{equation}
Let $J=[p,q]$. According to Lemma~\ref{lem:GC:intervals}, we have \citep[Theorem 1]{Adaptive:ICML:15}
\begin{equation} \label{eqn:UMA:inequality:5}
\sum_{i=-m}^n \sqrt{|I_i|} \leq 2 \sum_{i=0}^\infty \sqrt{2^{-i} |J|} \leq \frac{2\sqrt{2}}{\sqrt{2}-1} \sqrt{|J|} \leq 7 \sqrt{|J|}=7 \sqrt{q-p+1}.
\end{equation}
We get (\ref{eqn:UMA:inequality:2}) by combining (\ref{eqn:UMA:inequality:4}) and (\ref{eqn:UMA:inequality:5}).

When all the online functions are $\lambda$-strongly convex, Definition~\ref{def:strong} implies
\[
\begin{split}
&\sum_{t=p}^q f_t(\w_t) -  \sum_{t=p}^q f_t(\w) \\
\leq & \sum_{t=p}^q  \langle \nabla f_t(\w_t), \w_t -\w \rangle  - \frac{\lambda}{2} \sum_{t=p}^q \| \w_t -\w \|^2\\
\overset{\text{(\ref{eqn:UMA:inequality:1})}}{\leq} & 10 DG  \hat{a}(p,q) b(p,q) + 3 G\sqrt{\hat{a}(p,q) b(p,q)} \sqrt{\sum_{t=p}^q \| \w_t -\w \|^2}  -\frac{\lambda}{2} \sum_{t=p}^q \| \w_t -\w \|^2\\
\leq & \left(10 DG    + \frac{9G^2}{2\lambda} \right) \hat{a}(p,q) b(p,q).
\end{split}
\]

\subsection{Proof of Lemma~\ref{lem:meta:UMA}}
The analysis is similar to that of Lemma~\ref{lem:meta}.  We first demonstrate that (\ref{eqn:lem:meta:1}) also holds for the new surrogate loss $\hat{\ell}_t^\eta(\cdot)$.

Notice that
\begin{equation}\label{eqn:lem:meta:new:1}
\langle \nabla f_t(\w_t),  \w_t -\w \rangle^2 \leq \|\nabla f_t(\w_t)\|^2 \|\w_t -\w\|^2.
\end{equation}
As a result, we have
\begin{equation}\label{eqn:lem:meta:new:2}
\begin{split}
\exp\left(-\hat{\ell}_t^\eta(\w) \right) = &\exp\left(\eta \langle \nabla f_t(\w_t),  \w_t -\w \rangle - \eta^2 \|\nabla f_t(\w_t)\|^2\|\w_t -\w \|^2 \right) \\
\overset{\text{(\ref{eqn:lem:meta:new:1})}}{\leq} & \exp\left(\eta \langle \nabla f_t(\w_t),  \w_t -\w \rangle - \eta^2 \langle \nabla f_t(\w_t),  \w_t -\w \rangle^2 \right) =\exp\left(-\ell_t^\eta(\w) \right) \\
\overset{\text{(\ref{eqn:lem:meta:1})}}{\leq} & 1+ \eta \langle \nabla f_t(\w_t),  \w_t -\w \rangle
\end{split}
\end{equation}
for any $\w \in \Omega$.

Then, we repeat the derivation of (\ref{eqn:lem:meta:2}), and have
\[
\begin{split}
& \sum_{E_J^\eta \in \A_t} \exp(-L_{t,J}^\eta) + \sum_{\widehat{E}_J^\eta \in \widehat{\A}_t} \exp(-\widehat{L}_{t,J}^\eta)\\
=& \sum_{E_J^\eta \in \A_t} \exp(-L_{t-1,J}^\eta) \exp\left(-\ell_t^\eta(\w_{t,J}^\eta) \right) + \sum_{\widehat{E}_J^\eta \in \widehat{\A}_t} \exp(-\widehat{L}_{t-1,J}^\eta) \exp\left(-\hat{\ell}_t^\eta(\wh_{t,J}^\eta) \right)\\
\overset{\text{(\ref{eqn:lem:meta:1}),(\ref{eqn:lem:meta:new:2})}}{\leq} & \sum_{E_J^\eta \in \A_t} \exp(-L_{t-1,J}^\eta)\left(1+ \eta \langle \nabla f_t(\w_t),  \w_t -\w_{t,J}^\eta \rangle \right)\\
&+  \sum_{\widehat{E}_J^\eta \in \widehat{\A}_t} \exp(-\widehat{L}_{t-1,J}^\eta)\left(1+ \eta \langle \nabla f_t(\w_t),  \w_t -\wh_{t,J}^\eta \rangle \right)\\
\end{split}
\]
\[
\begin{split}
 =& \sum_{E_J^\eta \in \A_t} \exp(-L_{t-1,J}^\eta) +\sum_{\widehat{E}_J^\eta \in \widehat{\A}_t} \exp(-\widehat{L}_{t-1,J}^\eta) \\
 &+\left \langle \nabla f_t(\w_t),  \left(\sum_{E_J^\eta \in \A_t} \exp(-L_{t-1,J}^\eta) \eta +\sum_{\widehat{E}_J^\eta \in \widehat{\A}_t} \exp(-\widehat{L}_{t-1,J}^\eta) \eta\right)\w_t \right \rangle\\
 & - \left \langle \nabla f_t(\w_t), \sum_{E_J^\eta \in \A_t} \exp(-L_{t-1,J}^\eta) \eta \w_{t,J}^\eta + \sum_{\widehat{E}_J^\eta \in \widehat{\A}_t} \exp(-\widehat{L}_{t-1,J}^\eta) \eta \wh_{t,J}^\eta \right \rangle\\
\overset{\text{(\ref{eqn:wt:new})}}{=} & \sum_{E_J^\eta \in \A_t} \exp(-L_{t-1,J}^\eta) + \sum_{\widehat{E}_J^\eta \in \widehat{\A}_t} \exp(-\widehat{L}_{t-1,J}^\eta) .
\end{split}
\]
Following the derivation of (\ref{eqn:lem:meta:3}) and (\ref{eqn:lem:meta:4}), we have
\[
\begin{split}
&\sum_{E_J^\eta \in \A_s} \exp(-L_{s,J}^\eta) + \sum_{t=1}^{s-1}  \sum_{E_J^\eta \in \A_{t} \setminus \A_{t+1} } \exp(-L_{t,J}^\eta)  \\
&+ \sum_{\widehat{E}_J^\eta \in \widehat{\A}_s} \exp(-\widehat{L}_{s,J}^\eta) + \sum_{t=1}^{s-1}  \sum_{\widehat{E}_J^\eta \in \widehat{\A}_{t} \setminus \widehat{\A}_{t+1} } \exp(-\widehat{L}_{t,J}^\eta)\\
\leq  & |\A_1| + \sum_{t=2}^{s}| \A_t \setminus \A_{t-1}| + |\widehat{\A}_1| + \sum_{t=2}^{s}| \widehat{\A}_t \setminus \widehat{\A}_{t-1}| \\
\leq &  2 s \left(\lfloor \log_2 s \rfloor+1\right) \left(1+ \left\lceil \frac{1}{2} \log_2 s \right \rceil\right) \leq 4s^2.
\end{split}
\]

Thus, for any interval $I=[r,s] \in \I$, we have
\[
\exp(-L_{s,I}^\eta) = \exp\left( - \sum_{t=r}^s \ell_t^\eta(\w_{t,I}^\eta) \right)\leq  4s^2 \textrm{ and } \exp(-\widehat{L}_{s,I}^\eta) = \exp\left( - \sum_{t=r}^s \hat{\ell}_t^\eta(\wh_{t,I}^\eta) \right)\leq  4s^2
\]
which completes the proof.
\subsection{Proof of Lemma~\ref{lem:second:UMA}}
First, (\ref{eqn:second:1}) can be established by combining Lemmas \ref{lem:meta:UMA} and \ref{lem:expert}, and following the proof of  Lemma~\ref{lem:second}. Next, we prove (\ref{eqn:second:2}) in a similar way.

From Lemma~\ref{lem:strong:convex} and the property of AOGD \citep{NIPS2007_3319}, we have the following expert-regret of $\widehat{E}_I^\eta$ \citep[Theorem 2]{AAAI:2020:Wang}.
\begin{lemma}\label{lem:expert:strong} Under Assumptions~\ref{ass:1} and \ref{ass:2}, for any interval $I=[r,s] \in \I$ and any $\eta \in \S(s-r+1)$, the expert-regret of $\widehat{E}_I^\eta$ satisfies
\[
\sum_{t=r}^s \hat{\ell}_t^\eta(\wh_{t,I}^\eta) - \sum_{t=r}^s \hat{\ell}_t^\eta(\w)  \leq 1 + \log (s-r+2), \ \forall \w \in \Omega.
\]
\end{lemma}
Combining the regret bound in Lemmas~\ref{lem:meta:UMA} and \ref{lem:expert:strong}, we have
\[
\begin{split}
- \sum_{t=r}^s \hat{\ell}_t^\eta(\w) = & \eta \sum_{t=r}^s \langle \nabla f_t(\w_t),  \w_t -\w \rangle - \eta^2 \|f_t(\w_t)\|^2 \sum_{t=r}^s \|\w_t -\w \|^2 \\
\leq & 1+ 2\log_2(2s) + \log (s-r+2)
\end{split}
\]
for any $\eta \in \S(s-r+1)$. Thus,
\begin{equation} \label{eqn:lem:uma:1}
\begin{split}
\sum_{t=r}^s\langle \nabla f_t(\w_t),  \w_t -\w \rangle \leq &\frac{ 1+ 2\log_2(2s) + \log (s-r+2)}{\eta} + \eta \|\nabla f_t(\w_t)\|^2 \sum_{t=r}^s \|\w_t -\w \|^2 \\
\overset{\text{(\ref{eqn:gradient})}}{\leq} & \frac{ 1+ 2\log_2(2s) + \log (s-r+2)}{\eta} + \eta G^2 \sum_{t=r}^s \|\w_t -\w \|^2
\end{split}
\end{equation}
for any $\eta \in \S(s-r+1)$.

Let $\hat{a}(r,s)=1+ 2\log_2(2s) + \log (s-r+2) \geq 3$. Note that the optimal $\eta_*$ that minimizes the R.H.S.~of (\ref{eqn:lem:second:1}) is
\[
\eta_*=\sqrt{\frac{\hat{a}(r,s)}{G^2 \sum_{t=r}^s \|\w_t -\w \|^2}} \geq \frac{\sqrt{3}}{GD \sqrt{s-r+1}}.
\]
Recall that
\[
\S(s-r+1)= \left\{ \frac{2^{-i}}{5 DG} \ \left |\ i=0,1,\ldots, \left\lceil \frac{1}{2} \log_2 (s-r+1)\right\rceil \right. \right\}.
\]
If $\eta_* \leq \frac{1}{5DG}$, there must exist an $\eta \in \S(s-r+1)$ such that
\[
\eta \leq \eta_* \leq 2\eta.
\]
Then, (\ref{eqn:lem:uma:1}) implies
\begin{equation} \label{eqn:lem:uma:2}
\sum_{t=r}^s\langle \nabla f_t(\w_t),  \w_t -\w \rangle \leq  2\frac{ \hat{a}(r,s)}{\eta_*} + \eta_* G^2\sum_{t=r}^s \|  \w_t -\w \|^2 = 3 G\sqrt{\hat{a}(r,s)\sum_{t=r}^s \|\w_t -\w \|^2 }.
\end{equation}
On the other hand, if $\eta_* \geq \frac{1}{5DG}$,  we have
\[
\sum_{t=r}^s \|  \w_t -\w \|^2 \leq  25 D^2  \hat{a}(r,s).
\]
Then, (\ref{eqn:lem:uma:1}) with $\eta=\frac{1}{5DG}$ implies
\begin{equation} \label{eqn:lem:uma:3}
\sum_{t=r}^s\langle \nabla f_t(\w_t),  \w_t -\w \rangle \leq  5DG \hat{a}(r,s)+ 5 DG \hat{a}(r,s)   = 10 DG \hat{a}(r,s).
\end{equation}

We obtain (\ref{eqn:second:2}) by combining (\ref{eqn:lem:uma:2}) and (\ref{eqn:lem:uma:3}).

\section{Conclusion and future work}
In this paper, we develop a universal algorithm that is able to minimize the adaptive regret of general convex, exp-concave and strongly convex functions simultaneously. For each type of functions, our theoretical guarantee matches the performance of existing algorithms  specifically designed for this type of function under apriori knowledge of parameters.

In the literature, it is well-known that smoothness can be exploited to improve the static regret for different types of loss functions.
Recent studies \citep{jun2017,Adaptive:Regret:Smooth:ICML} have demonstrated that smoothness can be exploited to improve the adaptive regret, in analogy to the way that smoothness helps tighten the static regret \citep{NIPS2010_Smooth}. It is  an open question that whether our universal algorithm for minimizing the adaptive regret can be extended to support smoothness, and we will investigate it in the future.

\bibliography{E:/MyPaper/ref}

\begin{thebibliography}{34}
\providecommand{\natexlab}[1]{#1}
\providecommand{\url}[1]{\texttt{#1}}
\expandafter\ifx\csname urlstyle\endcsname\relax
  \providecommand{\doi}[1]{doi: #1}\else
  \providecommand{\doi}{doi: \begingroup \urlstyle{rm}\Url}\fi

\bibitem[Abernethy et~al.(2008)Abernethy, Bartlett, Rakhlin, and
  Tewari]{Minimax:Online}
Jacob Abernethy, Peter~L. Bartlett, Alexander Rakhlin, and Ambuj Tewari.
\newblock Optimal stragies and minimax lower bounds for online convex games.
\newblock In \emph{Proceedings of the 21st Annual Conference on Learning
  Theory}, pages 415--423, 2008.

\bibitem[Adamskiy et~al.(2012)Adamskiy, Koolen, Chernov, and
  Vovk]{Adamskiy2012}
Dmitry Adamskiy, Wouter~M. Koolen, Alexey Chernov, and Vladimir Vovk.
\newblock A closer look at adaptive regret.
\newblock In \emph{Proceedings of the 23rd International Conference on
  Algorithmic Learning Theory}, pages 290--304, 2012.

\bibitem[Arora et~al.(2012)Arora, Hazan, and Kale]{v008a006}
Sanjeev Arora, Elad Hazan, and Satyen Kale.
\newblock The multiplicative weights update method: a meta-algorithm and
  applications.
\newblock \emph{Theory of Computing}, 8\penalty0 (6):\penalty0 121--164, 2012.

\bibitem[Bartlett et~al.(2008)Bartlett, Hazan, and Rakhlin]{NIPS2007_3319}
Peter~L. Bartlett, Elad Hazan, and Alexander Rakhlin.
\newblock Adaptive online gradient descent.
\newblock In \emph{Advances in Neural Information Processing Systems 20}, pages
  65--72, 2008.

\bibitem[Boyd and Vandenberghe(2004)]{Convex-Optimization}
Stephen Boyd and Lieven Vandenberghe.
\newblock \emph{Convex Optimization}.
\newblock Cambridge University Press, 2004.

\bibitem[Cesa-Bianchi and Lugosi(2006)]{bianchi-2006-prediction}
Nicol\`{o} Cesa-Bianchi and G{\'a}bor Lugosi.
\newblock \emph{Prediction, Learning, and Games}.
\newblock Cambridge University Press, 2006.

\bibitem[Cesa-Bianchi et~al.(2005)Cesa-Bianchi, Mansour, and
  Stoltz]{Cesa-Bianchi2005}
Nicol{\`o} Cesa-Bianchi, Yishay Mansour, and Gilles Stoltz.
\newblock Improved second-order bounds for prediction with expert advice.
\newblock In \emph{Proceedings of the 18th Annual Conference on Learning
  Theory}, pages 217--232, 2005.

\bibitem[Chernov and Vovk(2009)]{Expert_Evaluators}
Alexey Chernov and Vladimir Vovk.
\newblock Prediction with expert evaluators' advice.
\newblock In \emph{Proceedings of the 20th International Conference on
  Algorithmic Learning Theory}, pages 8--22, 2009.

\bibitem[Cutkosky(2020)]{pmlr-v119-cutkosky20a}
Ashok Cutkosky.
\newblock Parameter-free, dynamic, and strongly-adaptive online learning.
\newblock In \emph{Proceedings of the 37th International Conference on Machine
  Learning}, pages 2250--2259, 2020.

\bibitem[Daniely et~al.(2015)Daniely, Gonen, and
  Shalev-Shwartz]{Adaptive:ICML:15}
Amit Daniely, Alon Gonen, and Shai Shalev-Shwartz.
\newblock Strongly adaptive online learning.
\newblock In \emph{Proceedings of the 32nd International Conference on Machine
  Learning}, pages 1405--1411, 2015.

\bibitem[Do et~al.(2009)Do, Le, and Foo]{icml2009_033}
Chuong Do, Quoc Le, and Chuan-Sheng Foo.
\newblock Proximal regularization for online and batch learning.
\newblock In \emph{Proceedings of the 26th International Conference on Machine
  Learning}, pages 257--264, 2009.

\bibitem[Freund et~al.(1997)Freund, Schapire, Singer, and
  Warmuth]{Freund:1997:UCP}
Yoav Freund, Robert~E. Schapire, Yoram Singer, and Manfred~K. Warmuth.
\newblock Using and combining predictors that specialize.
\newblock In \emph{Proceedings of the 29th Annual ACM Symposium on Theory of
  Computing}, pages 334--343, 1997.

\bibitem[Gaillard et~al.(2014)Gaillard, Stoltz, and van
  Erven]{pmlr-v35-gaillard14}
Pierre Gaillard, Gilles Stoltz, and Tim van Erven.
\newblock A second-order bound with excess losses.
\newblock In \emph{Proceedings of the 27th Conference on Learning Theory},
  pages 176--196, 2014.

\bibitem[Gy\"{o}rgy et~al.(2012)Gy\"{o}rgy, Linder, and
  Lugosi]{Track_Large_Expert}
Andr\'{a}s Gy\"{o}rgy, Tam\'{a}s Linder, and G\'{a}bor Lugosi.
\newblock Efficient tracking of large classes of experts.
\newblock \emph{IEEE Transactions on Information Theory}, 58\penalty0
  (11):\penalty0 6709--6725, 2012.

\bibitem[Hazan(2016)]{Intro:Online:Convex}
Elad Hazan.
\newblock Introduction to online convex optimization.
\newblock \emph{Foundations and Trends in Optimization}, 2\penalty0
  (3-4):\penalty0 157--325, 2016.

\bibitem[Hazan and Seshadhri(2007)]{Adaptive:Hazan}
Elad Hazan and C.~Seshadhri.
\newblock Adaptive algorithms for online decision problems.
\newblock \emph{Electronic Colloquium on Computational Complexity}, 88, 2007.

\bibitem[Hazan and Seshadhri(2009)]{Hazan:2009:ELA}
Elad Hazan and C.~Seshadhri.
\newblock Efficient learning algorithms for changing environments.
\newblock In \emph{Proceedings of the 26th Annual International Conference on
  Machine Learning}, pages 393--400, 2009.

\bibitem[Hazan et~al.(2007)Hazan, Agarwal, and Kale]{ML:Hazan:2007}
Elad Hazan, Amit Agarwal, and Satyen Kale.
\newblock Logarithmic regret algorithms for online convex optimization.
\newblock \emph{Machine Learning}, 69\penalty0 (2-3):\penalty0 169--192, 2007.

\bibitem[Herbster and Warmuth(1998)]{Herbster1998}
Mark Herbster and Manfred~K. Warmuth.
\newblock Tracking the best expert.
\newblock \emph{Machine Learning}, 32\penalty0 (2):\penalty0 151--178, 1998.

\bibitem[Jun et~al.(2017{\natexlab{a}})Jun, Orabona, Wright, and
  Willett]{Improved:Strongly:Adaptive}
Kwang-Sung Jun, Francesco Orabona, Stephen Wright, and Rebecca Willett.
\newblock Improved strongly adaptive online learning using coin betting.
\newblock In \emph{Proceedings of the 20th International Conference on
  Artificial Intelligence and Statistics}, pages 943--951, 2017{\natexlab{a}}.

\bibitem[Jun et~al.(2017{\natexlab{b}})Jun, Orabona, Wright, and
  Willett]{jun2017}
Kwang-Sung Jun, Francesco Orabona, Stephen Wright, and Rebecca Willett.
\newblock Online learning for changing environments using coin betting.
\newblock \emph{Electronic Journal of Statistics}, 11\penalty0 (2):\penalty0
  5282--5310, 2017{\natexlab{b}}.

\bibitem[Littlestone and Warmuth(1994)]{LITTLESTONE1994212}
Nick Littlestone and Manfred~K. Warmuth.
\newblock The weighted majority algorithm.
\newblock \emph{Information and Computation}, 108\penalty0 (2):\penalty0
  212--261, 1994.

\bibitem[Luo and Schapire(2015)]{pmlr-v40-Luo15}
Haipeng Luo and Robert~E. Schapire.
\newblock Achieving all with no parameters: Adanormalhedge.
\newblock In \emph{Proceedings of the 28th Conference on Learning Theory},
  pages 1286--1304, 2015.

\bibitem[Mhammedi et~al.(2019)Mhammedi, Koolen, and
  Van~Erven]{pmlr-v99-mhammedi19a}
Zakaria Mhammedi, Wouter~M Koolen, and Tim Van~Erven.
\newblock Lipschitz adaptivity with multiple learning rates in online learning.
\newblock In \emph{Proceedings of the 32nd Conference on Learning Theory},
  pages 2490--2511, 2019.

\bibitem[Shalev-Shwartz(2011)]{Online:suvery}
Shai Shalev-Shwartz.
\newblock Online learning and online convex optimization.
\newblock \emph{Foundations and Trends in Machine Learning}, 4\penalty0
  (2):\penalty0 107--194, 2011.

\bibitem[Shalev-Shwartz et~al.(2007)Shalev-Shwartz, Singer, and
  Srebro]{ICML_Pegasos}
Shai Shalev-Shwartz, Yoram Singer, and Nathan Srebro.
\newblock Pegasos: primal estimated sub-gradient solver for {SVM}.
\newblock In \emph{Proceedings of the 24th International Conference on Machine
  Learning}, pages 807--814, 2007.

\bibitem[Srebro et~al.(2010)Srebro, Sridharan, and Tewari]{NIPS2010_Smooth}
Nathan Srebro, Karthik Sridharan, and Ambuj Tewari.
\newblock Smoothness, low-noise and fast rates.
\newblock In \emph{Advances in Neural Information Processing Systems 23}, pages
  2199--2207, 2010.

\bibitem[van Erven and Koolen(2016)]{NIPS2016_6268}
Tim van Erven and Wouter~M Koolen.
\newblock {MetaGrad}: Multiple learning rates in online learning.
\newblock In \emph{Advances in Neural Information Processing Systems 29}, pages
  3666--3674, 2016.

\bibitem[Wang et~al.(2018)Wang, Zhao, and Zhang]{Adaptive:One:Gradient}
Guanghui Wang, Dakuan Zhao, and Lijun Zhang.
\newblock Minimizing adaptive regret with one gradient per iteration.
\newblock In \emph{Proceedings of the 27th International Joint Conference on
  Artificial Intelligence}, pages 2762--2768, 2018.

\bibitem[Wang et~al.(2019)Wang, Lu, and Zhang]{Adaptive:Maler}
Guanghui Wang, Shiyin Lu, and Lijun Zhang.
\newblock Adaptivity and optimality: A universal algorithm for online convex
  optimization.
\newblock In \emph{Proceedings of the 35th Conference on Uncertainty in
  Artificial Intelligence}, 2019.

\bibitem[Wang et~al.(2020)Wang, Lu, Hu, and Zhang]{AAAI:2020:Wang}
Guanghui Wang, Shiyin Lu, Yao Hu, and Lijun Zhang.
\newblock Adapting to smoothness: A more universal algorithm for online convex
  optimization.
\newblock In \emph{Proceedings of the 34th AAAI Conference on Artificial
  Intelligence}, pages 6162--6169, 2020.

\bibitem[Zhang et~al.(2018)Zhang, Yang, Jin, and Zhou]{Dynamic:Regret:Adaptive}
Lijun Zhang, Tianbao Yang, Rong Jin, and Zhi-Hua Zhou.
\newblock Dynamic regret of strongly adaptive methods.
\newblock In \emph{Proceedings of the 35th International Conference on Machine
  Learning}, 2018.

\bibitem[Zhang et~al.(2019)Zhang, Liu, and Zhou]{Adaptive:Regret:Smooth:ICML}
Lijun Zhang, Tie-Yan Liu, and Zhi-Hua Zhou.
\newblock Adaptive regret of convex and smooth functions.
\newblock In \emph{Proceedings of the 36th International Conference on Machine
  Learning}, pages 7414--7423, 2019.

\bibitem[Zinkevich(2003)]{zinkevich-2003-online}
Martin Zinkevich.
\newblock Online convex programming and generalized infinitesimal gradient
  ascent.
\newblock In \emph{Proceedings of the 20th International Conference on Machine
  Learning}, pages 928--936, 2003.

\end{thebibliography}

\appendix
\section{Additional Proofs}
For the sake of completeness, we provide the proofs of Lemmas \ref{lem:concave}, \ref{lem:strong:convex}, \ref{lem:expert} and \ref{lem:expert:strong}  \citep{Adaptive:Maler,AAAI:2020:Wang}.

\subsection{Proof of Lemma~\ref{lem:concave}}
Let $\nabla^2 \ell_t^{\eta}(\w)$ denote the Hessian matrix of $\ell_t^{\eta}(\cdot)$. By the definition of $\ell_t^{\eta}(\cdot)$ in \eqref{eqn:metagrad:3}, we have
\[
\begin{split}
&\nabla\ell_t^{\eta}(\w)\nabla\ell_t^{\eta}(\w)^{\top}\\
=&4\eta^4\nabla f_t(\w_t) \nabla f_t(\w_t)^{\top}(\w-\w_t)(\w-\w_t)^{\top}\nabla f_t(\w_t) \nabla f_t(\w_t)^{\top}\\
&+4\eta^3\nabla f_t(\w_t)(\w-\w_t)^{\top}\nabla f_t(\w_t) \nabla f_t(\w_t)^{\top}+\eta^2 \nabla f_t(\w_t)\nabla f_t(\w_t)^{\top}\\
=&\left(4\eta^3\langle \nabla f_t(\w_t),
\w-\w_t\rangle +4\eta^4\langle \nabla f_t(\w_t), \w-\w_t\rangle^2+\eta^2\right)\nabla f_t(\w_t)\nabla f_t(\w_t)^{\top}\\
\preceq &2\eta^2 \nabla f_t(\w_t)\nabla f_t(\w_t)^{\top}
=\nabla^2\ell _t^{\eta}(\w)
\end{split}
\]
for any $\w\in\Omega$, where the inequality is due to
$$4\eta^3\langle \nabla f_t(\w_t),
\w-\w_t\rangle +4\eta^4\langle \nabla f_t(\w_t), \w-\w_t\rangle^2 \overset{\text{(\ref{eqn:domain}),(\ref{eqn:gradient})}}{\leq} 4\eta^3GD+4\eta^4G^2D^2\leq \eta^2.$$
Thus, according to Lemma 4.1 of \citet{Intro:Online:Convex}, $\ell_t^{\eta}(\cdot)$ is 1-exp-concave.

Next, we upper bound the gradient of $\ell_t^{\eta}(\cdot)$ as follows:
\[
\begin{split}
\max\limits_{\w\in\Omega}\|\nabla \ell_t^{\eta}(\w)\|= &\max\limits_{\w\in\Omega} \left \| \eta  \nabla f_t(\w_t)    + 2 \eta^2 \langle \nabla f_t(\w_t),  \w -\w_t \rangle  \nabla f_t(\w_t) \right\|\\
\overset{\text{(\ref{eqn:domain}),(\ref{eqn:gradient})}}{\leq} & \eta G+2\eta^2G^2D\leq\frac{7}{25D}.
\end{split}
\]
\subsection{Proof of Lemma~\ref{lem:strong:convex}}
First, we show that
$$\hat{\ell}_t^{\eta}(\y)\geq \hat{\ell}_t^{\eta}(\x)+\langle \nabla \hat{\ell}_t^{\eta}(\x), \y-\x \rangle+\frac{2\eta^2\|\nabla f_t(\w_t)\|^2}{2}\|\y-\x\|^2$$
for any $\x,\y\in\Omega.$ When $\|\nabla f_t(\w_t)\|\not=0$, it is easy to verify that $\hat{\ell}_t^{\eta}(\cdot)$ is $2\eta^2\|\nabla f_t(\w_t)\|^2$-strongly convex, and the above inequality holds according to Definition \ref{def:strong}. When $\|\nabla f_t(\w_t)\|=0$, then by the definition of $\hat{\ell}_t^{\eta}(\cdot)$ in \eqref{eqn:ell:hat}, we have
\[
\hat{\ell}_t^{\eta}(\w)=\nabla \hat{\ell}_t^{\eta}(\w)=2\eta^2\|\nabla f_t(\w_t)\|^2=0
\]
 for any $\w\in\Omega$, and thus the inequality still holds.

Next, we upper bound the gradient of $\hat{\ell}_t^{\eta}(\cdot)$ as follows:
\begin{equation*}
\begin{split}
\|\nabla \hat{\ell}_t^{\eta}(\w)\|^2
=&\left\langle \eta \nabla f_t(\w_t)+2\eta^2\|\nabla f_t(\w_t)\|^2(\w-\w_t), \eta \nabla f_t(\w_t)+2\eta^2\|\nabla f_t(\w_t)\|^2(\w-\w_t) \right\rangle\\
=&\eta^2\|\nabla f_t(\w_t)\|^2+4\eta^3\|\nabla f_t(\w_t)\|^2\langle \nabla f_t(\w_t), \w-\w_t \rangle +4\eta^4 \|\nabla f_t(\w_t)\|^4\|\w-\w_t\|^2\\
\overset{\text{(\ref{eqn:domain}),(\ref{eqn:gradient})}}{\leq} &\eta^2\|\nabla f_t(\w_t)\|^2+\frac{4}{5}\eta^2\|\nabla f_t(\w_t)\|^2+\frac{4}{25}\eta^2\|\nabla f_t(\w_t)\|^2\\
\leq &2\eta^2\|\nabla f_t(\w_t)\|^2.
\end{split}
\end{equation*}
\subsection{Proof of Lemma~\ref{lem:expert}}
The proof is similar to that of Theorem 2 in \citet{ML:Hazan:2007}. For any $\u\in\mathbb{R}^d$, $A\in\mathbb{R}^{d\times d}$, let $\|\u\|^2_A$ denote $\u^{\top} A\u$. Based on Lemmas~\ref{lem:exp} and~\ref{lem:concave}, we have
\begin{equation}
\label{eqn:proof:lem:expert:1}
\ell^{\eta}_t(\w^{\eta}_{t,I})-\ell^{\eta}_t(\w)\leq \langle \nabla \ell^{\eta}_t(\w^{\eta}_{t,I}), \w^{\eta}_{t,I}-\w \rangle -\frac{\beta}{2}\langle \nabla \ell^{\eta}_t(\w^{\eta}_{t,I}), \w^{\eta}_{t,I}-\w \rangle^2
\end{equation}
for any $\w\in\Omega$, where $\beta=\frac{1}{2}\min\left( \frac{1}{4D \frac{7}{25D}}, 1 \right)=\frac{25}{56}$.
On the other hand, from the update rule in Algorithm \ref{alg:ONS}, we get
\begin{equation*}
\begin{split}
\|\w^{\eta}_{t+1,I}-\w\|^2_{\Sigma_t}\leq &\left\| \w_{t,I}^\eta - \frac{1}{\beta} \Sigma_t^{-1} \nabla \ell_t^\eta(\w_{t,I}^\eta)-\w\right\|_{\Sigma_t}^2\\
=& \| \w_{t,I}^\eta-\w\|^2_{\Sigma_{t}}-\frac{2}{\beta}\langle \nabla \ell_t^{\eta}(\w_{t,I}^\eta), \w_{t,I}^\eta-\w\rangle+\frac{1}{\beta^2}\|\nabla \ell_{t}^{\eta}(\w_{t,I}^\eta)\|_{\Sigma^{-1}_t}^2.
\end{split}
\end{equation*}
Based on the above inequality, we have
$$\langle\nabla \ell_t^{\eta}(\w_{t,I}^\eta),\w_{t,I}^\eta-\w\rangle \leq \frac{1}{2\beta}\|\nabla \ell_{t}^{\eta}(\w_{t,I}^\eta)\|_{\Sigma^{-1}_t}^2+\frac{\beta}{2}\| \w_{t,I}^\eta-\w\|^2_{\Sigma_{t}}-\frac{\beta}{2}\| \w_{t+1,I}^\eta-\w\|^2_{\Sigma_{t}}.$$
Summing up over $t=r$ to $s$, we get that
\begin{equation}\label{eqn:proof:lem:expert:2}
\begin{split}
&\sum_{t=r}^s\langle\nabla \ell_t^{\eta}(\w_{t,I}^\eta),\w_{t,I}^\eta-\w\rangle \\
\leq & \frac{1}{2\beta}\sum_{t=r}^s\|\nabla \ell_{t}^{\eta}(\w_{t,I}^\eta)\|_{\Sigma^{-1}_t}^2+\frac{\beta}{2}\| \w_{r,I}^\eta-\w\|^2_{\Sigma_r}\\
&+\frac{\beta}{2}\sum_{t=r+1}^s\|\w_{t,I}^\eta-\w\|^2_{\Sigma_t-\Sigma_{t-1}}
-\frac{\beta}{2}\|\w_{s+1,I}^\eta-\w\|^2_{\Sigma_s}\\
\leq &\frac{1}{2\beta}\sum_{t=r}^s\|\nabla \ell_{t}^{\eta}(\w_{t,I}^\eta)\|_{\Sigma^{-1}_t}^2+\frac{\beta}{2}\sum_{t=r}^s\langle \nabla \ell_{t}^{\eta}(\w_{t,I}^\eta), \w_{t,I}^\eta-\w \rangle^2\\
&+\frac{\beta}{2} \|\w_{r,I}^\eta-\w\|^2_{\Sigma_r-\nabla \ell_{t}^{\eta}(\w_{r,I}^\eta)\nabla \ell_{t}^{\eta}(\w_{r,I}^\eta)^{\top}}\\
=&\frac{1}{2\beta}\sum_{t=r}^s\|\nabla \ell_{t}^{\eta}(\w_{t,I}^\eta)\|_{\Sigma^{-1}_t}^2+\frac{\beta}{2}\sum_{t=r}^s\langle \nabla \ell_{t}^{\eta}(\w_{t,I}^\eta), \w_{t,I}^\eta-\w \rangle^2+\frac{1}{2\beta}.
\end{split}
\end{equation}
Combining \eqref{eqn:proof:lem:expert:1} and \eqref{eqn:proof:lem:expert:2}, we obtain
\begin{equation*}
\begin{split}
\sum_{t=r}^s \ell_t^\eta(\w_{t,I}^\eta) - \sum_{t=r}^s \ell_t^\eta(\w) \leq &
\sum_{t=r}^s\langle \nabla \ell^{\eta}_t(\w^{\eta}_{t,I}), \w^{\eta}_{t,I}-\w \rangle -\frac{\beta}{2}\sum_{t=r}^s\langle \nabla \ell^{\eta}_t(\w^{\eta}_{t,I}), \w^{\eta}_{t,I}-\w \rangle^2\\
\leq & \frac{1}{2\beta}\sum_{t=r}^s\|\nabla \ell_{t}^{\eta}(\w_{t,I}^\eta)\|_{\Sigma^{-1}_t}^2+\frac{1}{2\beta}\\
\leq & \frac{1}{2\beta}d\log (s-r+2) +\frac{1}{2\beta}\\
\leq & 5d\log (s-r+2)+5
\end{split}
\end{equation*}
where the third inequality is due to the following lemma
\citep[Lemma 11]{ML:Hazan:2007}.
\begin{lemma}
\label{lem:hazan:07}
For $t=1,\dots,T$, let $\u_t\in\R^d$ be a sequence of vectors such that for some $r>0$, $\|\u_t\|\leq r$. Define $V_t=\sum_{t=1}^T\u_t\u_t^{\top}+\epsilon I$. Then
$$\sum_{t=1}^T \|\u_t\|_{V_t^{-1}}^2\leq d\log\left(\frac{r^2T}{\epsilon}+1\right).$$
\end{lemma}
\subsection{Proof of Lemma~\ref{lem:expert:strong}}
Let $\wh_{t+1,I}^{\eta'}=\wh_{t,I}^{\eta}-\frac{1}{\alpha_t}\nabla \hat{\ell}_t^{\eta}(\wh_{t,I}^{{\eta}})$. By Lemma~\ref{lem:strong:convex}, we have
\begin{equation*}
\begin{split}
\hat{\ell}_t^{\eta}(\wh_{t,I}^{{\eta}})-\hat{\ell}_t^{\eta}(\w) &\leq \langle \nabla \hat{\ell}_t^{\eta}(\wh_{t,I}^{{\eta}}), \wh_{t,I}^{{\eta}}-\w \rangle -\frac{2\eta^2\|\nabla f_t(\w_t)\|^2}{2}\| \wh_{t,I}^{{\eta}}-\w\|^2\\
&= \alpha_t \langle \wh_{t,I}^{{\eta}}-\wh_{t+1,I}^{{\eta'}},\wh_{t,I}^{{\eta}}-\w\rangle -\frac{2\eta^2\|\nabla f_t(\w_t)\|^2}{2}\| \wh_{t,I}^{{\eta}}-\w\|^2
\end{split}
\end{equation*}
for any $\w\in\Omega$. For the first term, we have
\begin{equation*}
\begin{split}
&\langle \wh_{t,I}^{{\eta}}-\wh_{t+1,I}^{{\eta'}},\wh_{t,I}^{{\eta}}-\w\rangle\\
= & \| \wh_{t,I}^{{\eta}}-\w\|^2 + \langle \w-\wh_{t+1,I}^{{\eta'}}, \wh_{t,I}^{{\eta}}-\w \rangle\\
= & \| \wh_{t,I}^{{\eta}}-\w\|^2 - \| \wh_{t+1,I}^{{\eta'}}-\w\|^2 - \langle \wh_{t,I}^{{\eta}}-\wh_{t+1,I}^{{\eta'}}, \wh_{t+1,I}^{{\eta'}}-\w\rangle\\
=& \| \wh_{t,I}^{{\eta}}-\w\|^2 - \| \wh_{t+1,I}^{{\eta'}}-\w\|^2
+\| \wh_{t+1,I}^{{\eta'}}-\wh_{t,I}^{{\eta}}\|^2
 + \langle \wh_{t+1,I}^{{\eta'}}-\wh_{t,I}^{{\eta}}, \wh_{t,I}^{{\eta}}-\w\rangle
\end{split}
\end{equation*}
which implies that
\begin{equation*}
\begin{split}
 \langle \wh_{t,I}^{{\eta}}-\wh_{t+1,I}^{{\eta'}},\wh_{t,I}^{{\eta}}-\w\rangle=\frac{1}{2}
\left( \| \wh_{t,I}^{{\eta}}-\w\|^2 - \| \wh_{t+1,I}^{{\eta'}}-\w\|^2 + \| \wh_{t,I}^{{\eta}}-\wh_{t+1,I}^{{\eta'}}\|^2\right)
\end{split}
\end{equation*}
and thus
\begin{equation*}
\begin{split}
\hat{\ell}_t^{\eta}(\wh_{t,I}^{{\eta}})-\hat{\ell}_t^{\eta}(\w)
\leq &\frac{\alpha_t}{2}(\| \wh_{t,I}^{{\eta}} -\w\|^2 -\| \wh_{t+1,I}^{{\eta'}}-\w\|^2) \\
&+\frac{1}{2\alpha_t}\| \nabla \hat{\ell}^{\eta}_t(\wh_{t,I}^{{\eta}})\|^2-\frac{2\eta^2\|\nabla f_t(\w_t)\|^2}{2}\| \wh_{t,I}^{{\eta}}-\w\|^2.
\end{split}
\end{equation*}
Summing up over $t=r$ to $s$, we have
\begin{equation*}
\begin{split}
&\sum_{t=r}^s \hat{\ell}_t^{\eta}(\wh_{t,I}^{{\eta}})-\sum_{t=r}^s \hat{\ell}(\w)\\
\leq & \frac{\alpha_r}{2}\|\wh_{r,I}^{{\eta}}-\w\|^2+\sum_{t=r}^s\left(\alpha_t-\alpha_{t-1}-2\eta^2\|\nabla f_t(\w_t)\|^2\right)\frac{\|\wh_{t,I}^{{\eta}}-\w\|^2}{2}+\frac{1}{2}\sum_{t=r}^s \frac{1}{\alpha_t}\|\nabla \hat{\ell}_t^{\eta}(\wh_{r,I}^{{\eta}})\|^2\\
\leq& 1+\frac{1}{2}\sum_{t=r}^s \frac{1}{\alpha_t}\|\nabla \hat{\ell}_t^{\eta}(\wh_{r,I}^{{\eta}})\|^2
\leq  1+ \frac{1}{2}\sum_{t=r}^s\frac{\|\nabla f_t(\w_t)\|^2}{G^2+\sum_{i=r}^t\|\nabla f_t(\w_t)\|^2}\leq 1+\log (r-s+1)
\end{split}
\end{equation*}
where the second inequality is due to the fact that $\alpha_t-\alpha_{t-1}-2\eta^2\|\nabla f_t(\w_t)\|^2=0$ and $\eta\leq \frac{1}{5DG}$, the third inequality is derived from Lemma~\ref{lem:strong:convex}, and the last inequality is due to Lemma
~\ref{lem:hazan:07} (when $d=1$).

\end{document}